\documentclass[10pt,twocolumn,letterpaper]{article}

\usepackage{titlesec}
\usepackage{amsmath, amssymb}
\usepackage{algorithm}
\usepackage{algpseudocode}
\usepackage{mathtools}
\usepackage[dvipsnames]{xcolor}

\usepackage[pagenumbers]{cvpr}

\definecolor{cvprblue}{rgb}{0.21,0.49,0.74}
\usepackage[pagebackref,breaklinks,colorlinks,allcolors=cvprblue]{hyperref}

\def\paperID{10588} 
\def\confName{CVPR}
\def\confYear{2025}

\title{Reward Fine-Tuning Two-Step Diffusion Models via Learning Differentiable Latent-Space Surrogate Reward}

\author{Zhiwei Jia, Yuesong Nan, Huixi Zhao, Gengdai Liu \\
Zoom  Communications \\
{\tt\small \{firstname.lastname\}@zoom.com}
}

\begin{document}
\maketitle
\begin{abstract}

Recent research has shown that fine-tuning diffusion models (DMs) with arbitrary rewards, including non-differentiable ones, is feasible with reinforcement learning (RL) techniques, enabling flexible model alignment. 
However, applying existing RL methods to step-distilled DMs is challenging for ultra-fast ($\le2$-step) image generation.
Our analysis suggests several limitations of policy-based RL methods such as PPO or DPO toward this goal.
Based on the insights, we propose fine-tuning DMs with learned differentiable surrogate rewards.
Our method, named \textbf{LaSRO}, learns surrogate reward models in the latent space of SDXL to convert arbitrary rewards into differentiable ones for effective reward gradient guidance.
LaSRO leverages pre-trained latent DMs for reward modeling and tailors reward optimization for $\le2$-step image generation with efficient off-policy exploration.
LaSRO is effective and stable for improving ultra-fast image generation with different reward objectives, outperforming popular RL methods including DDPO \cite{black2023training} and Diffusion-DPO \cite{wallace2024diffusion}. 
We further show LaSRO's connection to value-based RL, providing theoretical insights.
See our webpage \href{https://sites.google.com/view/lasro}{here}.



\end{abstract}
    
\section{Introduction}
\label{sec:intro}

Diffusion models (DMs) \cite{sohl2015deep,song2019generative,song2020score,ho2020denoising} have significantly transformed generative modeling for continuous data, delivering remarkable results across various data modalities.
In particular, recent advances \cite{Ramesh2022HierarchicalTI,saharia2022photorealistic,Rombach2021HighResolutionIS} demonstrate their capabilities of generating diverse and high-fidelity images according to free-form input text prompts, albeit at the cost of a slow iterative sampling process.
Various methods are proposed to accelerate such a sampling process
\cite{liu2022flow,lipman2022flow,song2020denoising,lu2022dpm},
with the most recent distillation-based methods \cite{salimans2022progressive,liu2023instaflow,song2023consistency,sauer2023adversarial} pushing the envelope to achieve extreme few-step generation ($\le2$ steps).

On the other hand, state-of-the-art image diffusion models pre-trained on Internet-scale datasets (e.g. LAION-5B \cite{schuhmann2022laion}) usually incorporate a fine-tuning stage that guides the model’s behavior towards specific human preferences (aesthetics, debiased, etc.) 
This is achieved either by supervised learning (SFT) with curated human preference datasets \cite{xu2024imagereward,kirstain2023pick,wu2023human} or through reinforcement learning (RL) to leverage a diverse set of reward signals, including those not naturally differentiable such as incompressibility, or potentially the ones involving large language models (LLMs) or vision-language models (VLMs) \cite{huang2023t2i,lu2024llmscore,tan2024evalalign}.

However, it is non-trivial to fine-tune time-distilled DMs for $\le2$-step image generation with policy-based RL (e.g., PPO \cite{schulman2017proximal} or DPO \cite{rafailov2024direct}).
We detailedly analyze the emerged optimization challenges negligible for non-distilled DMs. 
Most RL objectives are \textit{ineffective} for $\le2$-step DMs due to the deterministic $2^{\text{nd}}$ sampling step.
The mapping of the distilled 2-step sampler from noisy images to clean ones (and thus to most reward signals) is \textit{highly non-smooth}, making policy gradient estimation challenging.
RL methods relying on the denoising diffusion loss (such as Diffusion-DPO \cite{wallace2024diffusion}) are incompatible with such a property, leading to blurred images.
Moreover, the 2-step mapping lacks enough stochasticity, posing a \textit{hard exploration} problem.
These conditions make most RL methods infeasible or ineffective for fine-tuning two-step DMs.

\begin{figure*}
  \centering
    \caption{
    Generated images of resolution $1024^2$ with $\le2$ steps via LCM-SSD-1B \cite{luo2023latent} (baseline) and those fine-tuned with Image Reward \cite{xu2024imagereward}. 
    $1^{\text{st}}$ column: baseline results.
    $2^{\text{nd}}$ and $3^{\text{rd}}$: results during and after fine-tuning the baseline via our method LaSRO.
    $4^{\text{th}}$: fine-tuned via RLCM \cite{oertell2024rl} (a variant of DDPO \cite{black2023training}).
    $5^{\text{th}}$: fine-tuned via PSO \cite{miao2024tuning} (a variant of Diffusion-DPO \cite{wallace2024diffusion}).
    Ours significantly improves the visual quality of $\le2$-step image generation while other strong RL methods fail due to training instability and inefficiency.}
    \label{fig:teaser}
    \includegraphics[width=\textwidth]{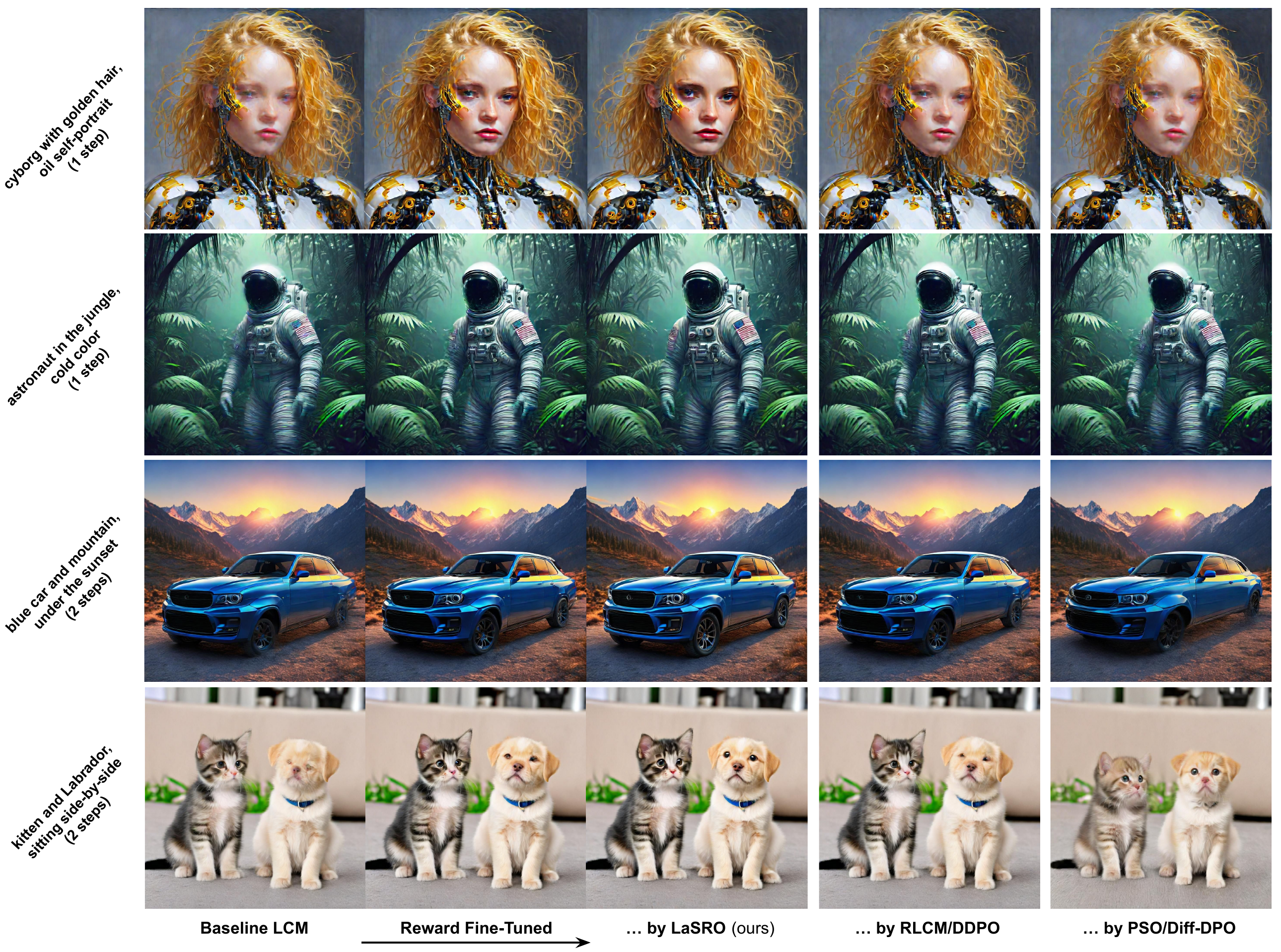}
\end{figure*}

To address these limitations, we propose to learn a surrogate reward model in the latent space (of SDXL \cite{podell2023sdxl}) that converts an arbitrary reward signal into a differentiable one.
Reward fine-tuning with this learned surrogate avoids policy gradient estimation as it directly and efficiently provides reward gradient guidance.
Specifically, we propose to use a pre-trained latent diffusion model as the backbone of the surrogate model, which shows superior memory \& speed efficiency and better generalizability compared to the alternatives based on VLMs (CLIP \cite{radford2021learning} and BLIP \cite{li2022blip}).
Additionally, our method achieves efficient exploration for 2-step image DMs with off-policy samples.
The overall pipeline consists of two stages: the pre-training stage for learning the surrogate reward and the reward optimization stage that alternates between improving the DMs and online adapting the surrogate.
We call our method \textbf{LaSRO}, i.e., \textbf{La}tent-space \textbf{S}urrogate \textbf{R}eward \textbf{O}ptimization.
LaSRO effectively fine-tunes two-step DMs across different reward objectives (including non-differentiable ones) and outperforms popular RL methods (DDPO \cite{black2023training}, Diffusion-DPO \cite{wallace2024diffusion}, etc.).
We further show a connection of LaSRO to value-based RL \cite{watkins1992q} to provide theoretical insights behind our approach.
In summary, our contributions are:
\begin{itemize}
    \item We analyze the key challenges in applying RL to improve $\le2$-step image diffusion models.
    \item We propose a framework, named LaSRO, that uses latent-space surrogate rewards to optimize arbitrary reward signals with efficient off-policy exploration.
    \item We establish LaSRO's connection to value-based RL, demonstrate its advantages over popular policy-based RL methods, and validate its design with ablation studies.
\end{itemize}




\section{Related Work}
\label{sec:related}

\paragraph{Diffusion Models (DMs)}

Diffusion models \cite{sohl2015deep,song2019generative,song2020score,ho2020denoising} have become the dominant generative models for various continuous data modalities \cite{ramesh2021zero,rombach2022high,ho2022video,Ho2022ImagenVH,kong2020diffwave,liu2023audioldm,Poole2022DreamFusionTU,Zeng2022LIONLP,janner2022planning,chi2023diffusion}.
Particularly, recent advances \cite{Ramesh2022HierarchicalTI,saharia2022photorealistic,Rombach2021HighResolutionIS} make text-to-image generation vastly popular.
A major drawback of DMs is their lengthy iterative denoising process during inference, often requiring many sampling steps.

\paragraph{Step-Distilled Diffusion Models}

Recent research strives to achieve $\le 2$ step generation through step-distillation of a teacher diffusion model.
Latent Consistency Models (LCMs) \cite{luo2023latent} apply Consistency Distillation \cite{song2023consistency} to learn to map any point along the sampling trajectories of the teachers (latent DMs) to their endpoints with one sampling step.
Progressive Distillation \cite{salimans2022progressive} incrementally distills a diffusion model into a faster one.
InstaFlow \cite{liu2023instaflow} apply distillation after modifying the teacher DM's probability flow to be smoother. 
Adversarial Diffusion Distillation \cite{sauer2023adversarial} distills a teacher into a faster student via adversarial training.



\paragraph{Fine-tuning DMs with Arbitrary Reward Signals}

Reward fine-tuning DMs with RL becomes promising for aligning model outputs with arbitrary reward signals, including non-differentiable ones.
The simplest form of RL comes in as reward-weighted regression (RWR) \cite{peters2007reinforcement}, including \cite{dong2023raft,huang2023t2i,lee2023aligning}.
Others \cite{wallace2024diffusion,yang2024using,deng2024prdp} based on direct preference optimization, DPO \cite{rafailov2024direct}, or its variant \cite{liang2024step},  can be seen as RWR-based approaches that fall into this category.
Policy-based RL methods \cite{fan2023optimizing,black2023training,fan2024reinforcement,fan2023dpok,chen2024find} relying on policy gradient or its equivalence handle arbitrary rewards by using large quantities of online samples.
Value-based RL methods \cite{watkins1992q} have been largely under-explored for fine-tuning DMs since they can struggle with long-horizon MDPs with delayed rewards \cite{sutton2018reinforcement}.
On the other hand, supervised fine-tuning methods \cite{xu2024imagereward,clark2023directly,prabhudesai2023aligning,wu2024deep} align DMs by back-propagating the reward gradient through the denoising process to update the model parameters.
While effective, they are constrained to scenarios where specifically labeled preference datasets are available \cite{Wu2023BetterAT,wu2023human,kirstain2023pick,xu2024imagereward} or the reward signal is differentiable. 
Many recent rewards \cite{huang2023t2i,lu2024llmscore,tan2024evalalign} are not naturally differentiable as they involve VLMs \cite{li2022blip} or LLMs \cite{achiam2023gpt}.

\paragraph{Reward Fine-tuning $\le2$-Step Step-Distilled DMs}

Fine-tuning DMs with policy-based RL remains a great challenge for $\le2$-step generation.
The distilled diffusion sampler lacks stochasticity and the underlying mapping is highly non-smooth, making methods based on either policy gradient estimation or reward-weighted regression ineffective (see Sec. \ref{sec:limit} for details). 
Very recently, RLCM \cite{oertell2024rl} and PSO \cite{miao2024tuning} were proposed to reward fine-tuning step-distilled DMs with policy-based RL, and RG-LCD \cite{li2024reward} was proposed to use reward guidance in the step-distillation process of LCMs.
However, they are ineffective or inefficient in improving $\le 2$-step DMs with arbitrary rewards.



\section{Preliminaries} \label{sec:prelim}


\subsection{Diffusion Models (DMs)}

We consider the conditional denoising diffusion probabilistic models \cite{sohl2015deep, ho2020denoising}, which represent an image generation distribution \( p(\mathbf{x}_0 | \mathbf{c}) \) over a sample image \( \mathbf{x}_0 \) given a text prompt \( \mathbf{c} \). 
The distribution is modeled as the reverse of a Markovian diffusion process \( q(\mathbf{x}_{t+1} | \mathbf{x}_{t}) \) that progressively adds Gaussian noise $Z_t$ to (a scaled version of) $\mathbf{x}_0$. 
To reverse the diffusion, a time-conditioned neural network, parametrized with $\epsilon-$prediction as \( \epsilon_\theta(\mathbf{x}_t, t, \mathbf{c}) \), is trained by maximizing a variational lower bound on the log-likelihood of the training data, or
equivalently, minimizing the loss:
\begin{equation} \label{eq:ddpm_loss}
    \mathcal{L}_{\text{ddpm}}(\theta) = \mathbb{E}\left[ \| Z_t - \epsilon_\theta(\mathbf{x}_t, t, \mathbf{c}) \|^2 \right]
\end{equation}
where the expectation is taken over all training samples $(\mathbf{x}_0, \mathbf{c})$, timestep $t \in \{0, ..., T\}$, and the corresponding noisy samples $\mathbf{x}_t \sim q(\mathbf{x}_t | \mathbf{x}_0)$ with the added noise $Z_t$. 
For simplicity, we ignore the scaling terms and denote the predicted output (rather than the noise) as $\mu_{\theta}(\mathbf{x}_t, t, \mathbf{c})$.

Sampling from a diffusion model starts with Gaussian noise \( \mathbf{x}_T \sim \mathcal{N}(\mathbf{0}, \mathbf{I}) \) and iteratively generates \(\mathbf{x}_{T-1}, \mathbf{x}_{T-2} \) all the way to \( \mathbf{x}_0 \). 
With some pre-set monotonically decreasing noise schedules $\{\sigma_t^2\}$, a common choice to parametrize the reverse denoising step given a timestep $t$ is:
\begin{equation} \label{eq:ddpm_step}
    p_\theta(\mathbf{x}_{t-1} | \mathbf{x}_t, \mathbf{c}) = \mathcal{N}(\mathbf{x}_{t-1} \mid \mu_\theta(\mathbf{x}_t, t, \mathbf{c}), \sigma_t^2 \mathbf{I})
\end{equation}

\subsection{Latent Consistency Models (LCMs)} \label{sec:lcm}

Aimed for fast image generation, the latent consistency model (LCM) \cite{luo2023latent}, a type of step-distilled DMs, was proposed to perform Consistency Distillation \cite{song2023consistency} in the latent space of the teacher DMs (e.g., SDXL \cite{podell2023sdxl}).
We select LCMs as our test bed due to their simplicity, efficiency, and training stability (our approach is \textit{not} tied to LCMs, see Tab. \ref{tab:numerical}).
For simplicity, we abuse the notation to use $\mathbf{x}_t$ to also refer to the latent code.
LCM learns a function $f_{\theta}(\mathbf{x}_{t}, t, \mathbf{c})$ that maps any sample $\mathbf{x}_{t}$ on a denoising trajectory to the endpoint $\mathbf{x}_0$ by ``compressing'' the trajectory induced by the teacher DMs (see Fig. \ref{fig:mix} upper right).
It enforces its output given $\mathbf{x}_{t}$ to be close to its output given $\mathbf{x}_{t-1}$, while overriding $f_{\theta}(\mathbf{x}_0, 0, \mathbf{c}) = \mathbf{x}_0$ for the boundary condition.
The (simplified) LCM training loss is:
\begin{equation} \label{eq:lcm_loss}
    \mathcal{L}_{\text{lcm}}(\theta) = \mathbb{E} \left[ d \left( f_\theta(\mathbf{x}_{t}, t, \mathbf{c}), f_{\theta}(\hat{\mathbf{x}}_{t-1}^\phi, t-1, \mathbf{c}) \right) \right]
\end{equation}
where \( d(\cdot, \cdot) \) is a metric function; the expectation is taken over $\mathbf{x}, \mathbf{c}$ and $t$;
stop gradient is applied to the $2^{\text{nd}}$ $f_{\theta}$ in $d(\cdot, \cdot)$; \( \hat{\mathbf{x}}_{t-1}^\phi \) is a sample obtained from \( \mathbf{x}_{t} \) with one denoising step using the teacher diffusion model $\phi$ given $\mathbf{c}$.

While LCMs produce images with only one step, by re-injecting noise back to the results of each LCM step (except for the last one),
LCMs benefit from a multi-step sampling procedure similar to regular DMs.
Namely, similar to Eq. \ref{eq:ddpm_step}, each LCM sampling step (except for the last one) follows:
\begin{equation} \label{eq:cm_step}
    p_\theta(\mathbf{x}_{t-1} | \mathbf{x}_t, \mathbf{c}) = \mathcal{N}(\mathbf{x}_{t-1} \mid f_{\theta}(\mathbf{x}_t, t, \mathbf{c}), \sigma_t^2 \mathbf{I})
\end{equation}
Again, scaling terms are ignored. 
Usually, only up to 8 steps are taken for LCMs (much fewer than regular DMs).

\subsection{RL Formulation of Fine-tuning DMs \& LCMs} \label{sec:mdp}

We formulate RL fine-tuning of DMs using Markov Decision Process (MDP) \cite{sutton2018reinforcement}. 
Similar to \cite{black2023training}, we map a diffusion model to the following finite-horizon MDP \((\mathcal{S}, \mathcal{A}, P, R, \mu, H) \) with policy $\pi(\mathbf{a}_t | \mathbf{s}_t)$ and a task reward \( r(s, \mathbf{c}) \) that evaluates the image given the text prompt $\mathbf{c}$: 
\begin{align*}
\mathbf{s}_t \triangleq &(\mathbf{x}_{\tau_t}, \tau_t, \mathbf{c}) \in \mathcal{S}
\quad \mathbf{a}_t \triangleq \mathbf{x}_{\tau_{t+1}} \in \mathcal{A} \\
&\pi(\mathbf{a}_t | \mathbf{s}_t) \triangleq p_\theta (\mathbf{x}_{\tau_{t+1}} | \mathbf{x}_{\tau_t}, \mathbf{c}) \\
 P(&\mathbf{s}_{t+1} | \mathbf{s}_t, \mathbf{a}_t) \triangleq (\delta_{\mathbf{x}_{\tau_{t+1}}}, \delta_{\tau_{t+1}}, \delta_\mathbf{c}) \\
R(&\mathbf{s}_t, \mathbf{a}_t) = \begin{cases} 
      r(\mathbf{a}_t, \mathbf{c}) & \text{if } t = H - 1 \\
      0 & \text{otherwise} 
   \end{cases} \\
\mu &\triangleq (p(\mathbf{c}),\delta_{\tau_0}, \mathcal{N}(\mathbf{0}, \mathbf{I}))
\end{align*}
where \( \delta_y \) is the Dirac $\delta$-distribution and $p_\theta (\mathbf{x}_{\tau_{t+1}} | \mathbf{x}_{\tau_t}, \mathbf{c})$ is defined according to Eq. \ref{eq:ddpm_step} or \ref{eq:cm_step}.
Note that, in MDP formulation (and for the rest of this paper), in contrast to the previous notation, the initial timestep is denoted $\tau_0$, and after $H$ denoising steps, the last timestep is denoted $\tau_H$.

For LCMs ($f_{\theta}$ from Sec. \ref{sec:lcm}), since the intermediate sampling result lies in the image/latent space where a non-zero reward is defined, the MDP formulation differs by: 
\begin{equation} \label{eq:reward}
    R(\mathbf{s}_t, \mathbf{a}_t) = r(f_\theta(\mathbf{x}_{\tau_t}, \tau_t, \mathbf{c}), \mathbf{c}), \ \forall t
\end{equation}
\section{What are the RL Obstacles?} \label{sec:limit}

We present an insightful analysis of three major challenges.

\subsection{Hard Exploration for Two-Step Image DMs}

\begin{figure}
  \centering
  \begin{subfigure}{0.665\linewidth}
    \includegraphics[width=\textwidth]{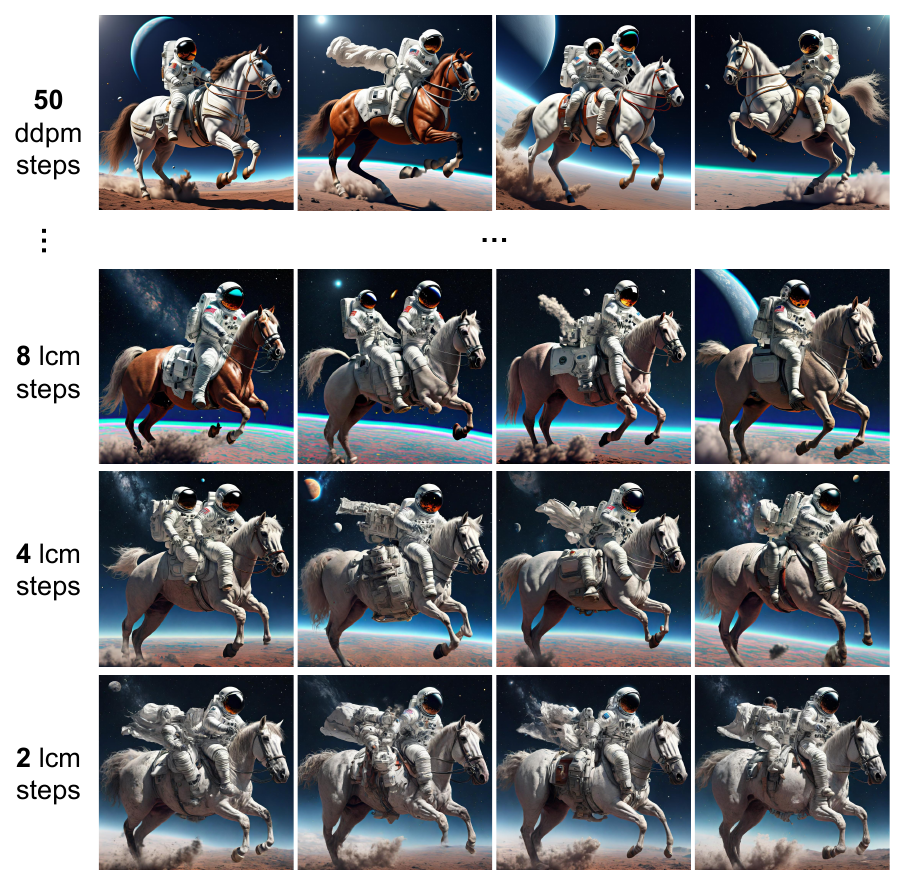}
  \end{subfigure}
  \hfill
  \begin{subfigure}{0.325\linewidth}
    \includegraphics[width=\textwidth]{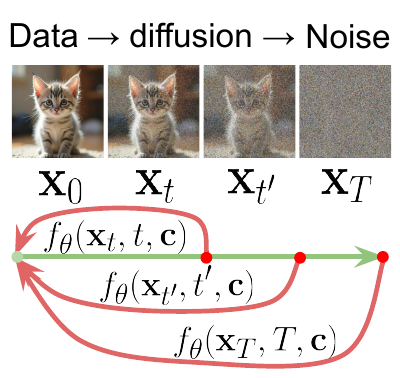} \\
    \vfill
    \includegraphics[width=\textwidth]{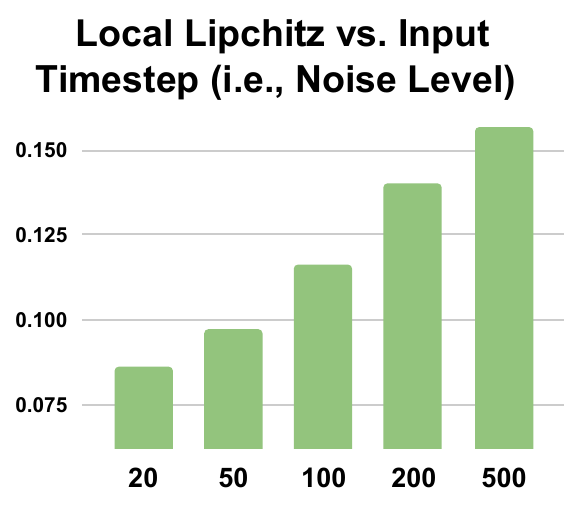}
  \end{subfigure}
  \caption{
  (\textbf{left}) Given \textit{astronaut riding a horse} as $\mathbf{c}$, fixed initial noise $\mathbf{x}_{\tau_0}$ and guidance scale, the reduction of noise-injecting steps leads to less diverse images generated from SSD-1B \cite{gupta2024progressive} and its LCM, and thus harder exploration for $p_\theta(\mathbf{x}_{\tau_H} | \mathbf{x}_{\tau_0}, \mathbf{c})$.
  (\textbf{right}) Illustration of the LCM mapping $f_{\theta}$.
  We show the growing empirical local Lipchitz of the LCM mapping from a noisy image to a clean image and to a generic image quality score (details in Appendix \ref{app:lipchitz}).
  The x-axis is the input's noise level ($t$ as in $\mathbf{x}_t$).}
  \label{fig:mix}
\end{figure}

Efficient and effective exploration in the state space is key to all online RL methods.
Fine-tuning two-step DMs poses a hard exploration problem for RL.
Diffusion models, expressed as stochastic policies $p_\theta(\mathbf{x}_{\tau_H} | \mathbf{x}_{\tau_0}, \mathbf{c})$, explore the image/latent space by repeatedly injecting Gaussian noise during the sampling process (Eq. \ref{eq:ddpm_step} or \ref{eq:cm_step}).
While regular DMs incrementally add noise to render diverse images, step-distilled DMs add noise in significantly fewer increments, leading to less diverse images, an observation we visualize in Fig. \ref{fig:mix} (left).
Given a fixed initial noise $\mathbf{x}_{\tau_0}$ and a prompt $\mathbf{c}$, one-step DMs (i.e., LCMs) are \textit{deterministic} functions,
and, by \textit{injecting noise only once}, two-step LCMs allow \textit{little exploration}:
\begin{equation} \label{eq:two_step_lcm}
     \mathbf{x}_{\tau_H} \sim f_{\theta}( f_{\theta}(\mathbf{x}_{\tau_0}, \tau_0, \mathbf{c}) + Z, \tau_{H/2}, \mathbf{c})  
\end{equation}
where scaling terms are ignored; $Z$ is the injected noise at timestep $\tau_{H/2}$ as the only source of stochasticity.
This limited exploration hinders on-policy RL methods.

\subsection{Degenerated RL Objectives for Two-Step DMs} 

Typically, policy-based RL optimizes the log-likelihood of online samples to enhance the policy’s expected reward.
However, most policy-based methods have their objectives degenerated when fine-tuning two-step DMs, substantially impeding their effectiveness.
For instance, a simple form of DDPO \cite{black2023training} to fine-tune DMs
has its objective and gradient:
\begin{equation} \label{eq:ddpo_loss}
    \mathcal{J}_{\text{ddpo}}(\theta) = \mathbb{E}_{\mathbf{c} \sim p(\mathbf{c}), \, \mathbf{x}_{\tau_H} \sim p_\theta(\mathbf{x}_{\tau_H} \mid \mathbf{c})} \left[ r(\mathbf{x}_{\tau_H}, \mathbf{c}) \right]    
\end{equation}
\begin{equation} \label{eq:ddpo_gradient}
       \nabla_\theta \mathcal{J}_{\text{ddpo}} = \mathbb{E} \left[ \sum_{t=0}^{H-1} \nabla_\theta \log p_\theta(\mathbf{x}_{\tau_{t+1}} | \mathbf{x}_{\tau_t}, \mathbf{c}) \, r(\mathbf{x}_{\tau_H}, \mathbf{c}) \right] 
\end{equation}
For two-step LCMs, since the $2^{\text{nd}}$ step is deterministic, $p_{\theta}(\mathbf{x}_{\tau_H}|\mathbf{x}_{\tau_{{H/2}}}, \mathbf{c})$ does not exist; Eq. \ref{eq:ddpo_loss} degenerates to:
\begin{equation} \label{eq:degenerated_policy_gradient}
       \mathcal{J}_{\text{ddpo}}^{\text{deg}} = 
       \mathbb{E} \left[\nabla_\theta \log p_\theta(\mathbf{x}_{\tau_{H/2}} | \mathbf{x}_{\tau_0}, \mathbf{c}) \, r(\mathbf{x}_{\tau_H}, \mathbf{c}) \right] 
\end{equation}
This objective only accounts for the \textit{first half} of the sampling process, making DDPO ineffective.
Similar issues can be found in other policy-based methods such as DPO \cite{miao2024tuning}.

\subsection{Non-Smooth Mappings of Two-Step DMs} \label{sec:non-smoothness}

The non-smoothness of the mapping that underlies two-step DMs makes it challenging to fine-tune them with methods based on either policy gradient estimation or reward-weighted regression (RWR).
Specifically, with a distilled two-step diffusion sampler, the underlying mapping from a noisy image to a clean image is highly non-smooth (i.e., has a great Lipchitz constant).
Thus, slight changes in actions could drastically change most rewards.
The same holds for two-step latent DMs:
an LCM step \( f_{\theta}(\mathbf{x}_{\tau_t}, \tau_t, \mathbf{c}) \) denoising an input with a higher noise level (\( \tau_t \) closer to \( \tau_0 \)) requires greater model expressiveness, resulting in a higher (local) Lipschitz constant \cite{lin2024sdxl}.
Note that, the $2^{\text{nd}}$ step of a two-step LCM handles a very high noise level from $\mathbf{x}_{\tau_{H/2}}$, empirically verified in Fig. \ref{fig:mix} (right).

This non-smoothness makes it difficult to estimate policy gradients, leading to training instability of PPO \cite{schulman2017proximal}.
In fact, many online RL methods that rely on policy gradient theorem \cite{sutton1999policy} suffer from high variance gradient estimation due to the non-smoothness.
Previously, ``sharpness'', a closely related concept, was adopted in the study of robustness of supervised learning, generative modeling, and RL \cite{jia2020information,foret2020sharpness,jia2021semantically,lee2024plastic}.
In contrast, many-step diffusion sampling ``divides'' the high Lipschitz so that policy gradient estimation becomes much easier for each step.

Methods based on RWR \cite{peters2007reinforcement} also suffer severely from the non-smoothness.
RWR mitigates high-variance policy gradient estimation by instead performing re-weighted regression of online samples to favor actions of higher rewards.
Methods based on DPO \cite{rafailov2024direct} fall into this category.
So does PRDP \cite{deng2024prdp}, which applies RWR adaptive to the reward difference between good and bad samples.
While RWR utilizes a re-weighted variant of the vanilla diffusion loss (Eq. \ref{eq:ddpm_loss}), \cite{miao2024tuning} shows that using Eq. \ref{eq:ddpm_loss} to fine-tune step-distilled DMs 
leads to blurry images by destroying the mapping induced by the step-distillation process. 
We argue it is because the loss does not account for the substantially increased non-smoothness of step-distilled DMs (particularly two-step DMs) compared to many-step DMs. 
PSO \cite{miao2024tuning} utilizes additional constraints in response, yet is ineffective for two-step DMs.

\section{Clearing the RL Obstacles with LaSRO}

\begin{figure}
  \centering
    \includegraphics[width=0.47\textwidth]{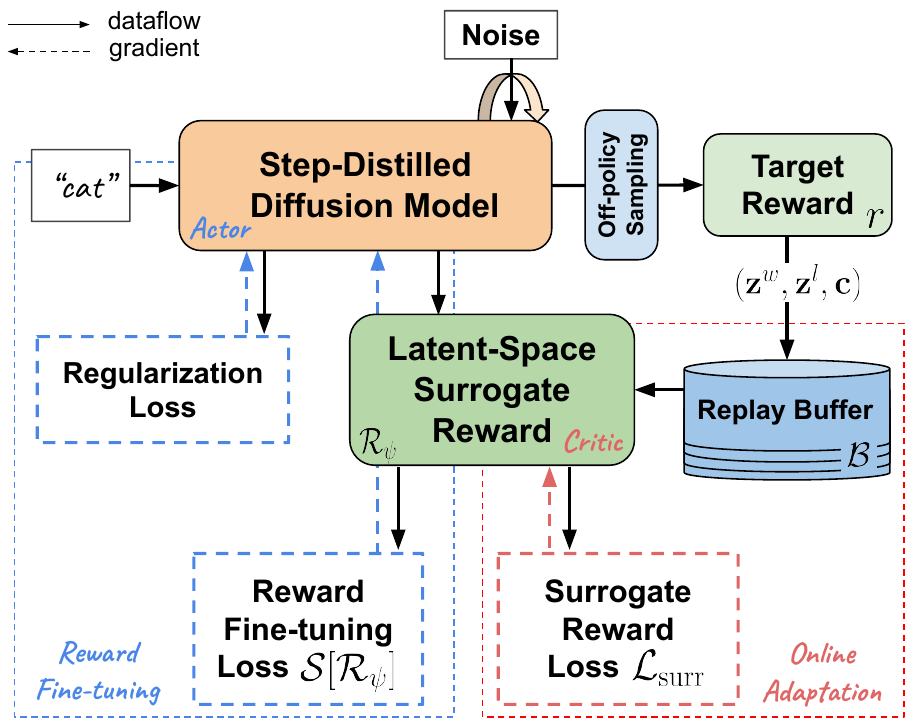}
  \caption{
  Training pipeline of the fine-tuning stage of LaSRO. This stage alternates between reward fine-tuning the DM and online adapting the surrogate reward. Since connected to value-based RL, we present LaSRO in the style of actor-critic methods \cite{konda1999actor}.}
  \label{fig:pipeline}
\end{figure}

In this paper, we focus on reward fine-tuning 2-step DMs, targeting ultra-fast image generation (please see App. \ref{app:impl} for the inference speed).
Formally, we introduce LaSRO, an RL training framework consisting of two stages: 
(1) pre-training the latent-space surrogate reward (Alg. \ref{algo:pretrain}).
(2) reward fine-tuning the two-step DMs (Alg. \ref{algo:ft}).
We visualize the pipeline of the latter stage in Fig. \ref{fig:pipeline}.

\subsection{Latent-space Surrogate Reward Learning} \label{sec:pretrain}

We start with surrogate reward learning.
Based on the insights from Sec. \ref{sec:limit},
we avoid both policy gradient estimation and reward-weighted regression.
Instead, we propose to learn a \textit{differentiable} surrogate reward that efficiently provides accurate reward gradient guidance for two-step image generation (guidance \textit{also} provided for the second deterministic step).
Direct reward gradient enabled by surrogate reward learning significantly improves exploration as noise injection is no longer the only way to explore when fine-tuning DMs.
Similar techniques (i.e., differentiable physics) were proposed in RL for robotics
\cite{hu2019difftaichi,huang2021plasticinelab,gillen2022leveraging}.

We leverage large pre-trained multimodal models to learn a surrogate reward that provides gradients generalizable over unseen prompts or images.
We evaluate different multimodal backbones including CLIP \cite{radford2021learning} (used as rewards in \cite{li2024reward}), BLIP \cite{li2022blip} (used as rewards in \cite{xu2024imagereward}), and the UNet encoder of SDXL \cite{podell2023sdxl} (used as discriminators in \cite{lin2024sdxl}).
We use the most effective one - the encoders of pre-trained latent DMs as the backbone.
We add a simple CNN-based prediction head on top of this backbone, together denoted as $\mathcal{R}_{\psi}$.
We train it with a loss based on the Bradley–Terry model \cite{bradley1952rank} in Eq. \ref{eq:pretrain} w.r.t. the target reward signal $r$ (potentially non-differentiable).
\begin{equation} \label{eq:pretrain}
     \mathcal{L}_{\text{surr}}(\psi; r) = - \mathbb{E}_{\mathbf{c}, \mathbf{z}^{w}, \mathbf{z}^{l}} \left[ \log( y_{\psi}^{\mathbf{c}}(\mathbf{z}^{w}, \mathbf{z}^{l})) \right]
\end{equation}
\begin{equation*}
\mathbf{z}^w \succeq \mathbf{z}^l \text{ ranked according to reward } r
\end{equation*}
\begin{equation*}
    y_{\psi}^{\mathbf{c}}(\mathbf{z}^{w}, \mathbf{z}^{l}) = \frac{\exp \left( \mathcal{R}_{\psi} (\mathbf{z}^{w}, \mathbf{c}) \right)}{\exp \left( \mathcal{R}_{\psi} (\mathbf{z}^{l}, \mathbf{c}) \right) + \exp \left( \mathcal{R}_{\psi} (\mathbf{z}^{w}, \mathbf{c}) \right)}
\end{equation*}
Namely, the surrogate reward $\mathcal{R}_{\psi}$ produces $\mathcal{R}_{\psi}(\mathbf{z}, \mathbf{c}) \in \mathbb{R}$ given prompt $\mathbf{c}$ and a latent code $\mathbf{z}$; 
the winning and losing pair $(\mathbf{z}^{w}, \mathbf{z}^{l})$ are selected from the samples (more on this shortly) according to the target reward $r$;
$\mathcal{L}_{\text{surr}}$ is a binary cross entropy loss that trains $\mathcal{R}_{\psi}$ to predict a higher score for the winning sample $\mathbf{z}^{w}$, thus transforming the knowledge from $r$ to $\mathcal{R}_{\psi}$.
Additional benefits of this latent-space reward are the \textit{improved computation \& memory efficiency} (no need to back-prop through VAEs) and the potential to \textit{capture image details better} ($\mathcal{R}_{\psi}$ operates in the learned downsampled latent space; others \cite{xu2024imagereward,kirstain2023pick,wu2023human} operates in the ``brute-force'' downsampled image space).

We target improvements in one and two-step image generation, a goal reflected in our design of \textit{both} stages.
In the pre-training stage with a pre-trained LCM $f_{\theta}$, in each iteration, given a randomly drawn prompt $\mathbf{c}$, we sample $N_s$ different $1^{\text{st}}$-step LCM outputs $\{\mathbf{z}_1 \triangleq f_{\theta}(\mathbf{z}_{\tau_0}, \tau_0, \mathbf{c})\}$ with $N_s$ initial noise $\{\mathbf{z}_{\tau_0}\}$ (\textit{off-policy} exploration, see Sec. \ref{sec:insight}) and then sample the $2^{\text{nd}}$ LCM step to get $\{\mathbf{z}_2 \triangleq f_{\theta}(\mathbf{z}_1 + Z, \tau_{H/2}, \mathbf{c})\}$ with injected noise $\{Z\}$ as in Eq. \ref{eq:two_step_lcm}.
Presented in Alg. \ref{algo:pretrain}, in the pre-training stage, we train $\mathcal{R}_{\psi}$ using $\{(\mathbf{z}_1^{w}, \mathbf{z}_1^{l}, \mathbf{c})\}$ selected from $\{\mathbf{z}_1\}$ and $\{(\mathbf{z}_2^{w}, \mathbf{z}_2^{l}, \mathbf{c})\}$ from $\{\mathbf{z}_2\}$, facilitating the downstream task - reward fine-tuning two-step DMs.

\begin{algorithm} 
\small
\caption{\small Pre-training Stage for Two-step DMs (e.g., LCMs)} \label{algo:pretrain}
\begin{algorithmic}[1] 
\Require prompt set $\mathcal{C}$; initial surrogate reward $\mathcal{R}_{\psi}$; LCM $f_{\theta}$; learning rate $\eta$; VAE decoder $\mathcal{D}$; target reward (can be non-differentiable) $r$; hyper-parameter $N_s$
\Repeat
    \State Sample prompt $\mathbf{c} \sim \mathcal{C}$
    \State Sample two-step LCM $f_{\theta}$ for $N_s$ times to get $\{\mathbf{z}_1\}$, $\{\mathbf{z}_2\}$ 

    \State Find a W/L pair $(\mathbf{z}_1^{w}, \mathbf{z}_1^{l})$ from $\{\mathbf{z}_1\}$ acc. to $r(\mathcal{D}(\mathbf{z}_1), \mathbf{c})$
    \State Find a W/L pair $(\mathbf{z}_2^{w}, \mathbf{z}_2^{l})$ from $\{\mathbf{z}_2\}$ acc. to $r(\mathcal{D}(\mathbf{z}_2), \mathbf{c})$

    \State Compute $\mathcal{L}_{\text{surr}}^1(\psi; r)$ from Eq. \ref{eq:pretrain} given $(\mathbf{z}_1^{w}, \mathbf{z}_1^{l}, \mathbf{c})$ 
    \State Compute $\mathcal{L}_{\text{surr}}^2(\psi; r)$ from Eq. \ref{eq:pretrain} given $(\mathbf{z}_2^{w}, \mathbf{z}_2^{l}, \mathbf{c})$ 
    
    \State $\psi \leftarrow \psi - \eta \nabla_\psi [ \mathcal{L}_{\text{surr}}^1(\psi; r) + \mathcal{L}_{\text{surr}}^2(\psi; r) ]$
\Until{convergence}
\end{algorithmic}
\end{algorithm}

\subsection{Direct Reward Fine-tuning of Two-Step DMs} \label{sec:ft_stage}

After pre-training, we enter the fine-tuning stage, which alternates between reward fine-tuning the DMs and online adapting the surrogate reward (which handles the distribution drift of the outputs of fine-tuned two-step DMs).
We present this stage in Fig. \ref{fig:pipeline} and in Alg. \ref{algo:ft}.

\begin{algorithm} 
\small
\caption{\small Fine-tuning Stage for Two-step DMs (e.g., LCMs)} \label{algo:ft}
\begin{algorithmic}[1] 
\Require (besides $\mathcal{C}$, $f_{\theta}$, $\mathcal{D}$, $r$, $N_s$ from Alg. \ref{algo:pretrain}) pre-trained surrogate $\mathcal{R}_{\psi}^*$; learning rates $\eta_1, \eta_2$; \# steps $N_1, N_2$; reg. loss $\mathcal{L}_{\text{lcm}}$ (Eq. \ref{eq:lcm_loss}); loss coeff. $c, c_1, c_2$; norm. and clipping function $\mathcal{S}$
\Repeat
    \State (Re-)initiate the replay buffer $\mathcal{B} = \emptyset$ 
    \Repeat \textcolor{RoyalBlue}{\quad \quad \quad \quad \quad \quad  \quad \quad \quad \quad ``Reward Fine-tuning''}        
        \State \textcolor{RoyalBlue}{Sample prompt $\mathbf{c} \sim \mathcal{C}$}
        \State \textcolor{RoyalBlue}{Sample $\{\mathbf{z}_1\}$, $\{\mathbf{z}_2\}$ as results of a two-step LCM $f_{\theta}$}

        \State \textcolor{RoyalBlue}{Compute $\mathcal{S}[\mathcal{R}_{\psi}(\mathbf{z}_1, \mathbf{c})]$ and $\mathcal{S}[\mathcal{R}_{\psi}(\mathbf{z}_2, \mathbf{c})]$}
        \State \textcolor{RoyalBlue}{Compute reg. loss with $\mathcal{L}_{\text{lcm}}$ and $\mathcal{L}_{\text{lasro}}$ from Eq. \ref{eq:ft}}
        \State \textcolor{RoyalBlue}{$\theta \leftarrow \theta - \eta_1 \mathcal{L}_{\text{lasro}}$}

        \State Find $\mathbf{z}_*^{w}, \mathbf{z}_*^{l}$ from $\{\mathbf{z}_1\}, \{\mathbf{z}_2\}$ with $r$, similar to Alg. \ref{algo:pretrain}
        \State Insert $(\mathbf{z}_1^{w}, \mathbf{z}_1^{l}, \mathbf{c})$ and $(\mathbf{z}_2^{w}, \mathbf{z}_2^{l}, \mathbf{c})$ to $\mathcal{B}$
    \Until{$N_1$ steps}
    \Repeat \textcolor{BrickRed}{\quad \quad \quad \quad \quad \quad  \quad \quad \quad \quad ``Online Adaptation''}
        \State \textcolor{BrickRed}{Sample $(\mathbf{z}_*^{w}, \mathbf{z}_*^{l}, \mathbf{c}) \sim \mathcal{B}$ to compute $\mathcal{L}_{\text{surr}}^*(\psi; r)$}
        \State \textcolor{BrickRed}{Update $\psi$ similar to Alg. \ref{algo:pretrain} with learning rate $\eta_2$}
    \Until{$N_2$ steps}
\Until{convergence}
\end{algorithmic}
\end{algorithm}

During reward fine-tuning, we use a normalized and clipped fine-tuning loss $\mathcal{S}[\mathcal{R}_{\psi}(\mathbf{z}_*, \mathbf{c})]$.
By keeping track of the moving average and ``maximum'' value of the surrogate reward, $\mathcal{S}$ essentially improves training stability (details in Appendix \ref{app:clipping}).
We use the original step-distillation loss for the distilled DMs (with a separate prompt and image set for distillation) as a regularization loss.
For LCM, it is $\mathcal{L}_{\text{lcm}}$ from Eq. \ref{eq:lcm_loss}.
The total fine-tuning loss is:
\begin{equation} \label{eq:ft}
    \mathcal{L}_{\text{lasro}} = c \mathcal{L}_{\text{lcm}} + c_1 \mathcal{S}[\mathcal{R}_{\psi}(\mathbf{z}_1, \mathbf{c})] + c_2 \mathcal{S}[\mathcal{R}_{\psi}(\mathbf{z}_2, \mathbf{c})]
\end{equation}
where $c, c_1, c_2$ are loss coefficients (hyper-parameters).

\subsection{Theoretical Insight - Value-Based RL} \label{sec:insight}

The superior performance of LaSRO over popular RL methods can be partly explained by its connection to value-based RL.
In contrast to pure policy-based RL that usually relies on on-policy samples to estimate policy gradients, value-based RL
\cite{watkins1992q,mnih2013playing} uses off-policy samples to learn action value functions $Q(\mathbf{s}_t, \mathbf{a}_t)$ that quantify how ``desirable'' a particular state and action is via temporal difference (TD) learning \cite{sutton1988learning}. 
Particularly for short-horizon MDPs with immediate rewards, value-based RL proves advantageous \cite{sutton2018reinforcement} because (1) the learned value function can reduce gradient noise and is more \textit{robust} than pure policy gradient methods and (2) a completely off-policy exploration strategy is much more \textit{efficient} than the alternatives.


Value-based methods are largely \textit{under-explored} for fine-tuning DMs.
For instance, existing methods like DDPO \cite{black2023training} utilize the clipped RL objective in PPO yet use no learned value function.
LaSRO shares several functional and structural similarities with value-based methods:
(\textbf{i}) The TD loss for horizon $H=2$ can be shown largely equivalent to the surrogate reward loss $\mathcal{L}_{\text{surr}}$ from Eq. \ref{eq:pretrain}, with the surrogate $\mathcal{R}_{\psi}$ largely equivalent to the value function $Q$.
(\textbf{ii}) Optimizing LCMs $f_{\theta}$ is guided by $\mathcal{R}_{\psi}$, similar to how policy is guided by value function.
(\textbf{iii}) Exploration is done in an efficient off-policy manner via also sampling the initial noise $\mathbf{x}_{\tau_0}$ for $N_s$ times, rather than restricted to the on-policy distribution $p_\theta (\mathbf{x}_{\tau_{H}}| \mathbf{x}_{\tau_0}, \mathbf{c})$ with a fixed $\mathbf{x}_{\tau_0}$.
This connection gives us insights into the superiority of LaSRO.
See more discussions in Appendix \ref{app:value-based-rl}.

\section{Experiments} \label{sec:exp}

We demonstrate the superior performance of LaSRO for fine-tuning two-step DMs over three distinctive rewards under the non-differentiable (black-box) reward setup \cite{uehara2024understanding}. 
Some rewards (e.g., the text-alignment score that involves sampling) are truly non-differentiable.
We further perform ablation studies to justify our design (training scheme, choice of architecture/backbone, etc.).

We choose LCMs \cite{luo2023latent} distilled from SSD-1B \cite{gupta2024progressive} as our major test bed since it is a strong baseline for 2-step image generation of $1024^2$ resolution (SDXL-Turbo \cite{sauer2023adversarial} focuses on $512^2$ images and SDXL-Lightning \cite{lin2024sdxl} does not support 1-step generation).
Our approach can also be applied to other step-distilled DMs, though (see Tab. \ref{tab:numerical}).

Please refer to App. \ref{app:impl} for implementation details.

\begin{figure}
  \centering
    \hspace{-5px}
  \begin{subfigure}{0.505\linewidth}
    \includegraphics[width=\textwidth]{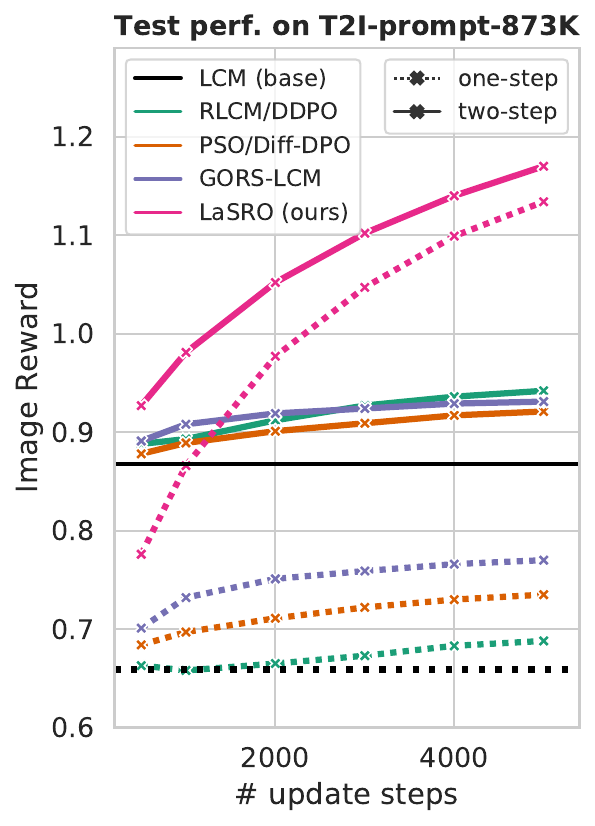}
  \end{subfigure}
    \hspace{-5px}
  \begin{subfigure}{0.485\linewidth}
    \includegraphics[width=1.02\textwidth]{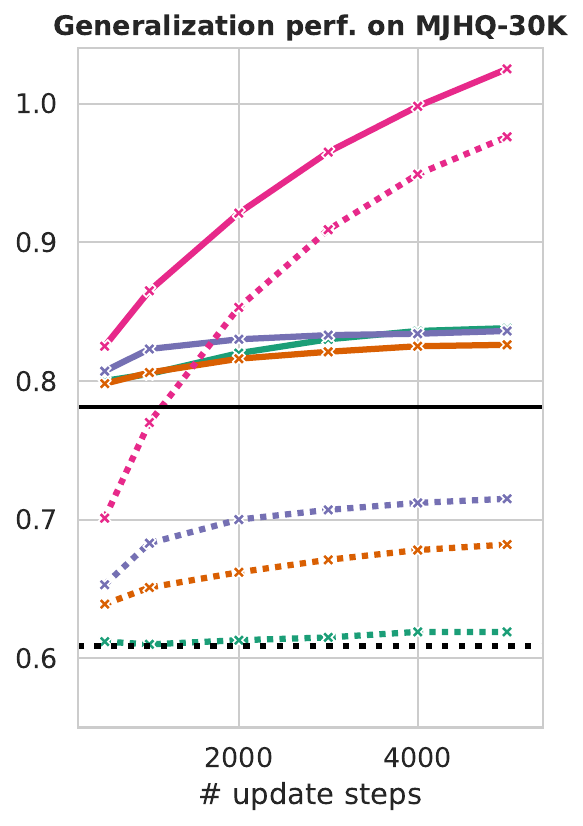}
  \end{subfigure}
  \caption{
   Comparison of LaSRO and other RL methods for fine-tuning two-step LCMs with Image Reward. Results on the test prompts of T2I-prompt-87K (\textbf{left}) and the (out-of-distribution) prompts of MJHQ-30K (\textbf{right}) show LaSRO effectively improve one-step (dashed lines) and two-step (solid) image generation.}
  \label{fig:plots_1}
\end{figure}

\subsection{Case Study - Improving General Image Quality} \label{sec:main_res1}

We apply LaSRO to improve the general image quality of two-step DMs by optimizing Image Reward \cite{xu2024imagereward}.
Image Reward is a popular metric for text-to-image evaluations by using a BLIP-based model (learned from labeled data) to rate images based on criteria such as text alignment and aesthetics.
It outperforms other metrics \cite{kirstain2023pick,wu2023human,schuhmann2022laion} in terms of alignment with human preference. 
We consider it a black box (while it is originally differentiable) and \textit{do not} use its gradient information for all compared methods.

\paragraph{Training Data}
We curate a set of 873k unique prompts, named T2I-prompt-873K, with text prompts from a combination of DiffusionDB \cite{wang2022diffusiondb} and open-prompts (\href{https://github.com/krea-ai/open-prompts}{link}).
We split it into 30k for the test split and the rest for training.

\paragraph{Evaluation Protocol} 
Besides the test set performance on T2I-prompt-873K, we also compare the generalization performance (both measured in Image Reward) over the prompts of MJHQ-30K \cite{li2024playground}, a set curated to balance 10 common text-to-image categories (thus, a different prompt distribution).
We tune all compared methods to mitigate reward overoptimization \cite{gao2023scaling} with the help of qualitative evaluation.
We report the improved Image Reward during training as curves and summarize each curve by selecting its best checkpoint according to the FID \cite{heusel2017gans} measured with the reference image set of MJHQ-30K, which accounts for the image fidelity (see details in Appendix \ref{app:overoptimize}).

\vspace{-5px}
\paragraph{Baselines and Results}
We compare LaSRO with strong RL fine-tuning baselines: GORS-LCM 
 (an iterative generate-filter-distill method adapted to LCMs based on GORS \cite{huang2023t2i}), PSO \cite{miao2024tuning} (a method adapted for step-distilled DMs based on DM-DPO \cite{wallace2024diffusion,yang2024using}), and RLCM \cite{oertell2024rl} (a variant of DDPO \cite{black2023training} for LCMs based on PPO \cite{schulman2017proximal}).
 Additionally, we apply LaSRO to fine-tuning SDXL-Turbo \cite{sauer2023adversarial}.
We report the results in Fig. \ref{fig:plots_1}, Tab. \ref{tab:numerical} and Fig. \ref{fig:teaser} \& \ref{fig:results}.

\begin{figure}
  \centering
  \begin{subfigure}{0.49\linewidth}
    \includegraphics[width=\textwidth]{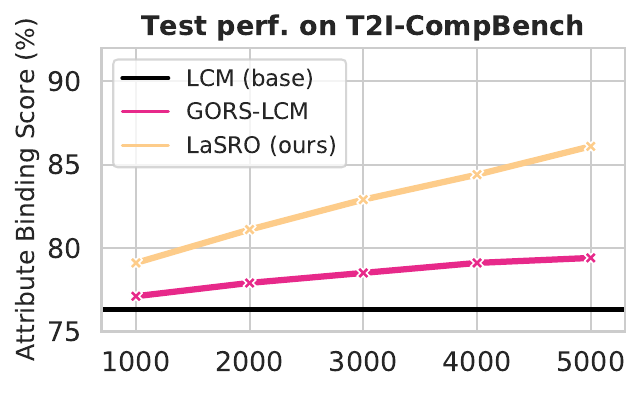} 
    \includegraphics[width=\textwidth]{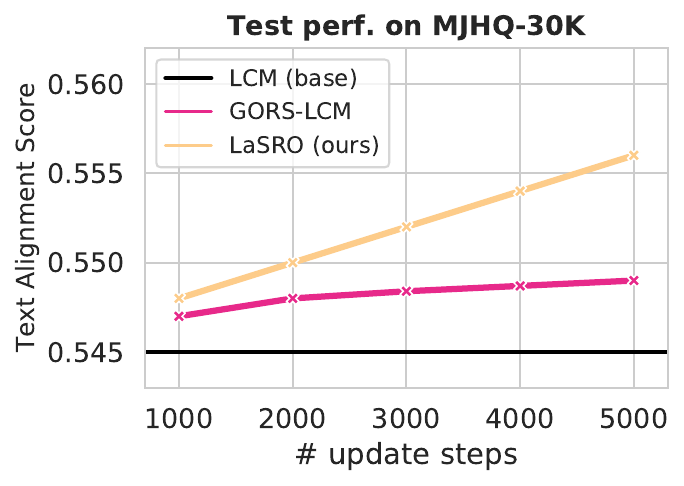}
  \end{subfigure}
  \hfill
  \begin{subfigure}{0.49\linewidth}
    \includegraphics[width=1.05\textwidth]{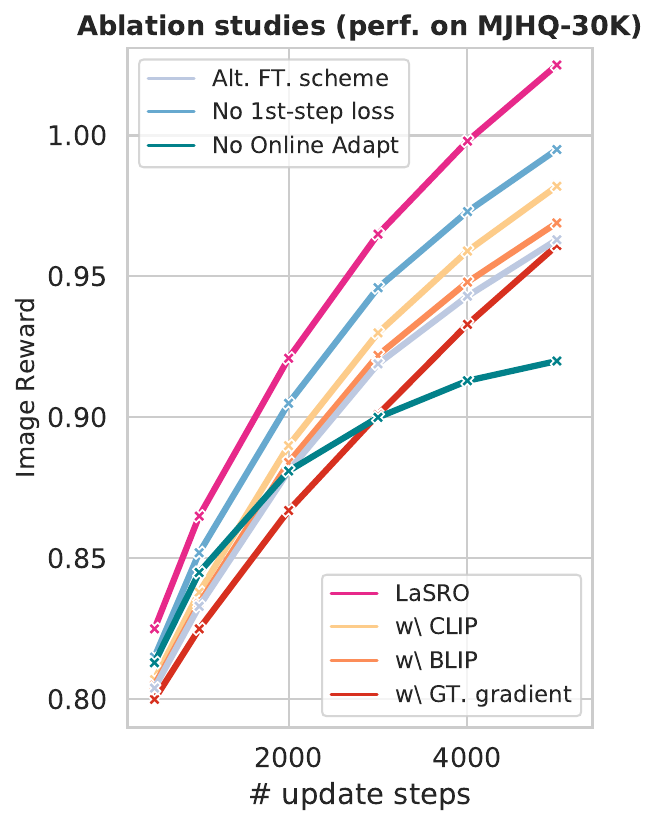}
  \end{subfigure}
  \caption{
  (\textbf{left}) LaSRO vs. GORS-LCM (other baselines fail) for fine-tuning LCMs with Attribute Binding Score and Text Alignment Score (all two-step perf. results).
  (\textbf{right}) Results from ablation studies that validate LaSRO's design (see Sec. \ref{sec:abl}).}
  \label{fig:plots_2}
\end{figure}

\begin{table}[h]
\centering
\resizebox{0.38\textwidth}{!}{%
\begin{tabular}{lccc}
\toprule
 & \# steps & resolution & Image Reward \\
 \midrule
\midrule
SDXL-Turbo \cite{sauer2023adversarial} & 1 & $512^2$ & 0.817 \\ 
SDXL-Turbo & 2 & $512^2$ & 0.839 \\
\ \ + LaSRO (ours) & 1 & $512^2$ & \textbf{0.904} \\
\ \ + LaSRO (ours) & 2 & $512^2$ & \textbf{0.957} \\
\midrule
SDXL-Lightning \cite{lin2024sdxl} & 2 & $1024^2$ & 0.634 \\
\midrule
SSD-1B-LCM \cite{luo2023latent} & 1 & $1024^2$ & 0.609 \\
SSD-1B-LCM & 2 & $1024^2$ & 0.781 \\
\ \ +RLCM/DDPO \cite{oertell2024rl,black2023training}& 2 & $1024^2$ & 0.830 \\
\ \ +PSO/Diff-DPO \cite{miao2024tuning,wallace2024diffusion} & 2 & $1024^2$ & 0.821 \\
\ \ + GORS-LCM \cite{huang2023t2i} & 2 & $1024^2$& 0.836 \\
\ \ + LaSRO (ours) & 1 & $1024^2$ & \textbf{0.946} \\
\ \ + LaSRO (ours) & 2 & $1024^2$ & \textbf{0.998} \\
\bottomrule
\end{tabular}%
}
\caption{Gen. perf. on MJHQ-30K of step-distilled DMs and the fine-tuned DMs with RL methods (see details in Appendix \ref{app:numerical}). 
We find that LaSRO is agnostic to step-distillation techniques.
}
\vspace{-5pt}
\label{tab:numerical}
\end{table}

\subsection{Additional Case Studies} \label{sec:main_res2}

Besides general image quality, we provide two case studies of applying LaSRO to optimize two-step DMs with \textit{non-differentiable} rewards designed for tailored criteria.

\begin{figure*}
  \centering
    \caption{Two-step results ($1024^2$ resolution) of LCM baseline and of fine-tuned LCM via LaSRO (with Image Reward for column 1-3 and with Attribute Binding Score for 4-5). LaSRO significantly improves both reward signals.
    See additional visual results in Appendix \ref{app:visual}.}
    \label{fig:results}
    \includegraphics[width=.95\textwidth]{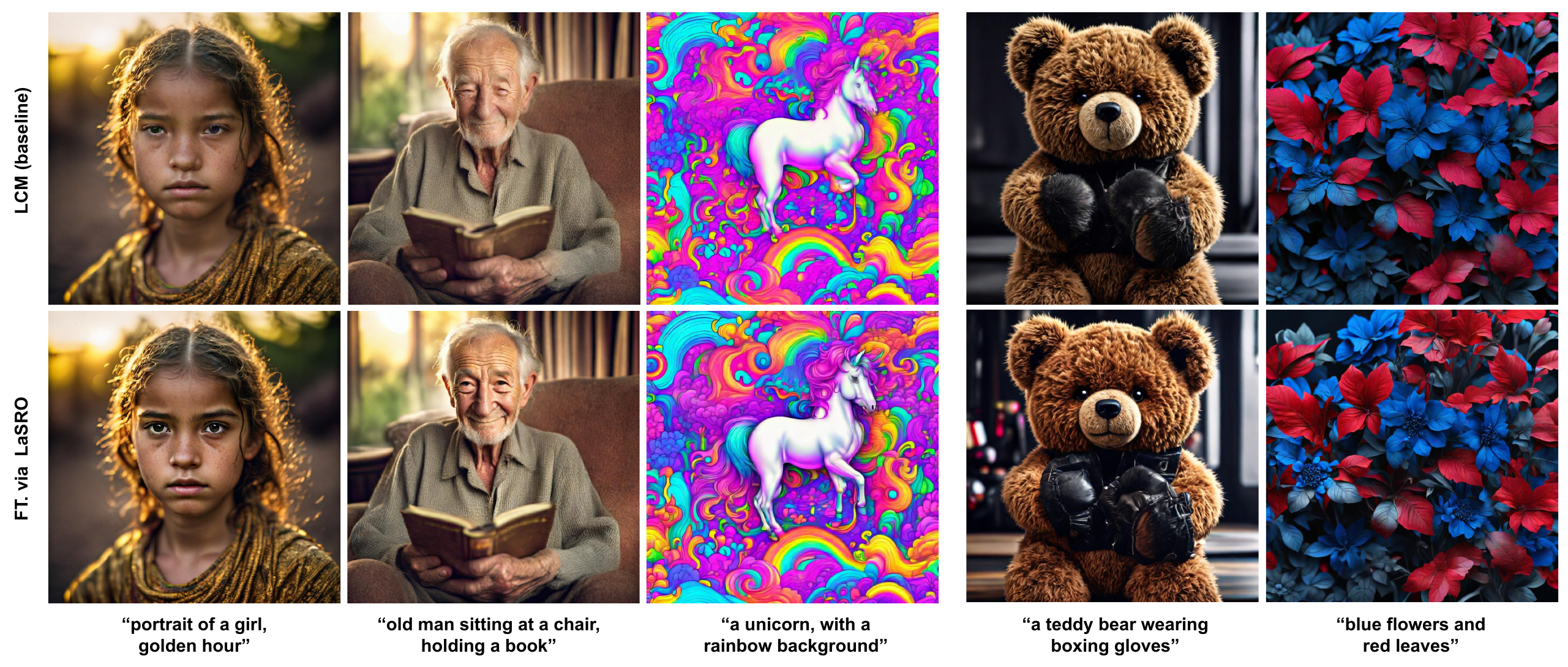}
\end{figure*}

\paragraph{Improving Attribute Binding}
we investigate using LaSRO to enhance attribute binding between the input prompts and the corresponding elements in generated images.
Inspired by T2I-CompBench \cite{huang2023t2i}, we derive a reward that uses BLIP-VQA \cite{li2022blip} to ask the generated image ``if the subject exists?'' and ``if it has the correct attribute?'' according to each noun phrase in the prompt (answering ``yes'' means score-1 and score-0 otherwise).
The reward then computes the average score among all questions across all (prompt, image) pairs. 
Denoted as \textit{Attribute Binding Score}, it is a non-differentiable reward reported as accuracy (\%) and is more intuitive than the original one from \cite{huang2023t2i} (which uses averaged raw output score for ``yes'').
It is also harder for RL fine-tuning as the reward is more ``discrete''.

\paragraph{Improving Text Alignment}
Inspired by \cite{black2023training}, we use a reward that first uses image captioning models (i.e., Florence-2 \cite{xiao2024florence}) to capture the detailed text description of the generated image and then uses BERTScore \cite{zhang2019bertscore} to compare the \textit{Text Alignment Score} between this description and the original text prompt.
This is a non-differentiable reward due to the image captioning procedure.

\paragraph{Training Data and Evaluation Protocol}
For the attribute binding task, we use the train split of the 3k attribute-binding prompts from T2I-CompBench for training and use the corresponding test split for evaluation and report the Attribute Binding Score.
Regarding text alignment, we set aside 27k prompts (train split) from MJHQ-30K for training and the rest (3k prompts) for evaluation with Text Alignment Score as the reported metric.

\paragraph{Baselines and Results}
We compare LaSRO to the LCM baseline and GORS-LCM as we find other RL baselines suffer from training instability and are outperformed by GORS-LCM in both tasks.
Please see results in Fig. \ref{fig:plots_2} (left).


\subsection{Ablation Studies} \label{sec:abl}
We conduct extensive ablation studies of LaSRO in improving the general image quality of two-step LCMs (by reporting the generalization perf. on MJHQ-30K).
(\textbf{a}) We use CLIP or BLIP (instead of the UNet of latent DMs) as the surrogate reward's backbone, denoted {\small \texttt{w\textbackslash CLIP}} and {\small \texttt{w\textbackslash BLIP}}. 
(\textbf{b}) We examine a variant {\small \texttt{Alt.FT.scheme}}, a reward fine-tuning scheme used in RG-LCM \cite{li2024reward} and widely adopted in supervised reward fine-tuning of ordinary DMs \cite{xu2024imagereward,clark2023directly}.
Compared to LaSRO, this scheme does not directly target $\le2$-step generation.
(\textbf{c}) We examine a variant {\small \texttt{No 1st-step loss}}, which set $c_1=0$ in Eq. \ref{eq:ft} to disable also optimizing the $1^{\text{st}}$-step LCM results.
(\textbf{d}) In {\small \texttt{No Online Adapt}}, we remove the surrogate reward online adaption substage in Alg. \ref{algo:ft}.
(\textbf{e}) We include a non-RL baseline that directly uses Image Reward to replace the surrogate reward (gradient info. from Image Reward is used), denoted {\small \texttt{w\textbackslash GT.gradient}}. 
We did not observe the high-frequency noise issue reported in \cite{li2024reward} that partly motivates their approach since, contrary to reward-guided distillation \cite{li2024reward}, we study fine-tuning after distillation.

The numerical results in Fig. \ref{fig:plots_2} (right) validate our design.
Additionally, we find LaSRO relatively robust against manual noise perturbation of the GT. reward, partly explaining why it outperforms GT. reward gradients.

\section{Conclusion} \label{sec:conclusion}


We address the challenges of fine-tuning 2-step image DMs with arbitrary reward signals.
Our analysis identifies limitations in applying existing RL methods.
In response, we propose LaSRO, a framework that learns differentiable surrogate rewards in the latent space with off-policy exploration to provide effective gradient guidance during DM fine-tuning.
We show LaSRO's superiority over existing RL methods for optimizing diverse reward signals.
We validate LaSRO's design through extensive ablation studies and further draw a connection between LaSRO and value-based RL to provide theoretical insights.
We believe this work opens the door for both large-scale RL fine-tuning and customized model alignment (through flexible reward design) of ultra-few-step image diffusion models, with the potential to be applied in other data modalities.
{
    \small
    \bibliographystyle{ieeenat_fullname}
    \bibliography{main}
}

\onecolumn

\setlength{\parindent}{0pt}  
\setlength{\parskip}{1em}   




\appendix

\clearpage
\setcounter{page}{1}


\section*{Appendix}

\section{Empirical Local Lipchitz of Two-step LCMs} \label{app:lipchitz}

\begin{figure}[h]
  \centering
  \begin{subfigure}{0.3\linewidth}
    \includegraphics[width=\textwidth]{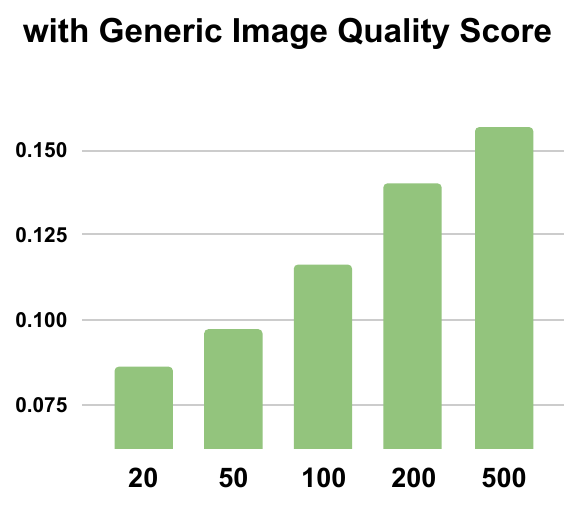}
  \end{subfigure}
  \begin{subfigure}{0.3\linewidth}
    \includegraphics[width=\textwidth]{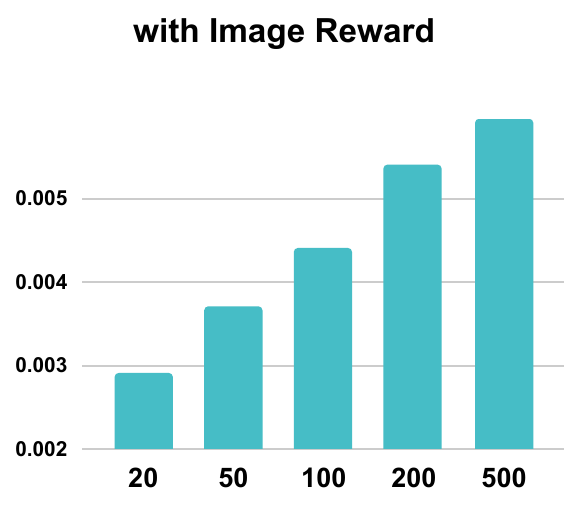}
  \end{subfigure}
  \caption{
  We show the growing empirical local Lipchitz of the mapping that underlies two-step LCM image generation. 
  The x-axis is the input data's noise level (i.e., timestep). The y-axis is the estimated local Lipchitz. 
  The left and right figures incorporate the reconstruction-based image quality score and the Image Reward, respectively, in estimating the local Lipchitz.}
  \label{fig:lipchitz}
\end{figure}

In Sec. \ref{sec:non-smoothness}, we argue that the non-smoothness of the mapping that underlies two-step DMs makes it challenging to fine-tune them.
Specifically, the mapping for which we measure the non-smoothness refers to the $2^{\text{nd}}$ step of a distilled two-step diffusion sampler (e.g., an LCM $f_{\theta}$) that maps a noisy image to a clean image. 
We present empirical evidence by estimating two different local Lipchitz constants characterizing this mapping. 

Formally, given a prompt $\mathbf{c}$ and initial noise $\mathbf{z}_{\tau_0}$, the $1^{\text{st}}$ LCM step (with re-injected noise) gives you $\mathbf{z}_1^{t} = f_{\theta}(\mathbf{z}_{\tau_0}, \tau_0, \mathbf{c}) + Z_{t}$, with notations of the denoised latent code $\mathbf{z}_1$ and the injected noise $Z_{t}$ at noise level (i.e., timestep) $t$ from Sec. \ref{sec:pretrain}.
Note that, in a standard two-step LCM inference setup ($H=2$), the noise $Z$ is injected at noise level $t = \tau_{H/2}=\tau_1$ (the middle of the denoising trajectory), and so $\mathbf{z}_1^t = \mathbf{z}_1^{\tau_{H/2}} =\mathbf{z}_1^{\tau_1}$.
Here, we use the notation that the injected noise level $t$ might not be $\tau_{H/2}$.
We then randomly sample another valid action $\mathbf{z}_1^{t}({\epsilon})$ ($\epsilon > 0$) in the neighborhood of $\mathbf{z}_1^{t}$, at the same noise level $t$, by $\mathbf{z}_1^{t}(\epsilon) = \sqrt{1 - \epsilon^2} \mathbf{z}_1^{t} + \epsilon \mathbf{z}'$, where $\mathbf{z}' \sim \mathcal{N}(0, \mathcal{I})$.
We set $\epsilon=0.01$ so that $\mathbf{z}_1^{t}({\epsilon})$ is close enough to $\mathbf{z}_1^{t}$ for the Lipchitz estimation.
Next, we obtain the $2^{\text{nd}}$ LCM results $\mathbf{z}_2^{t} = f_{\theta}(\mathbf{z}_1^{t}, t, \mathbf{c})$ and $\mathbf{z}_2^{t}(\epsilon) = f_{\theta}(\mathbf{z}_1^{t}(\epsilon), t, \mathbf{c})$.
To efficiently compute the local Lipchitz, we need a real-value function that further maps $\mathbf{z}_2^{t}$ and $\mathbf{z}_2^{t}(\epsilon)$ to scalars.
We use two functions, both of which help us to illustrate the non-smoothness property well.

We first propose a prompt-free generic image quality metric: the reconstruction error $\ell_{\text{rec}}(\mathbf{x}) = \| \mathcal{D}(\mathcal{E}(\mathbf{x})) - \mathbf{x} \|$ given a pre-trained VAE's encoder $\mathcal{E}$ and decoder $\mathcal{D}$ (we use VAE from SDXL \cite{podell2023sdxl}), which estimates how well a generated image $\mathbf{x}$ fit in the image manifold learned by the pre-trained VAE.
We use the Image Reward \cite{xu2024imagereward}, denoted $r(\mathbf{x},\mathbf{c})$ as the second function.
Namely, for $\ell_{\text{rec}}$ we have the following two terms:
\[
\ell_{\text{rec}}(\mathbf{z}_2^{t}) = \| \mathcal{D}(\mathcal{E}(\mathcal{D}(\mathbf{z}_2^{t}))) - \mathcal{D}(\mathbf{z}_2^{t}) \|
\]
\[
\ell_{\text{rec}}(\mathbf{z}_2^{t}(\epsilon)) = \| \mathcal{D}(\mathcal{E}(\mathcal{D}(\mathbf{z}_2^{t}(\epsilon)))) - \mathcal{D}(\mathbf{z}_2^{t}(\epsilon)) \|
\]
and for $r$ we have:
\[
r(\mathbf{z}_2^{t}, \mathbf{c}) = r(\mathcal{D}(\mathbf{z}_2^{t}), \mathbf{c})
\]
\[
r(\mathbf{z}_2^{t}(\epsilon), \mathbf{c}) = r(\mathcal{D}(\mathbf{z}_2^{t}(\epsilon)), \mathbf{c})
\]
Combined, we can then compute the two empirical local Lipschitz constants given the input data's noise level $t$ as:
\[
L_{\text{local}}(\ell_{\text{rec}}, t) = \mathbb{E}_{(\mathbf{z}_{\tau_0}, \mathbf{c})} \left[ \frac{|\ell_{\text{rec}}(\mathbf{z}_2^{t}) - \ell_{\text{rec}}(\mathbf{z}_2^{t}(\epsilon))|}{\|\mathbf{z}_1^{t} -  \mathbf{z}_1^{t}(\epsilon)\|} \right]
\]
\[
L_{\text{local}}(r, t) = \mathbb{E}_{(\mathbf{z}_{\tau_0}, \mathbf{c})} \left[ \frac{| r(\mathbf{z}_2^{t}, \mathbf{c}) - r(\mathbf{z}_2^{t}(\epsilon), \mathbf{c}) |}{\|\mathbf{z}_1^{t} -  \mathbf{z}_1^{t}(\epsilon)\|} \right] 
\]
Both empirical Lipchitz are estimated over $N=1000$ sampled $(\mathbf{z}_{\tau_0}, \mathbf{c})$ pairs.
To visualize the growing non-smoothness of the mapping when the input data has a larger noise level (i.e., the LCM covers a longer denoising trajectory in one single sampling step), we compute these two quantities with varying input noise level $t=20, 50, 100, 200, 500$. 

We illustrate the numerical values of these two local Lipchitz for a pre-trained LCM based on SSD-1B \cite{gupta2024progressive} in Fig. \ref{fig:lipchitz}.

\section{Normalization and Clipping Function} \label{app:clipping}

Similar to the advantage estimation \cite{schulman2015high} and the clipped RL objective proposed in PPO \cite{schulman2017proximal}, we utilize a normalization and clipping function $\mathcal{S}$ to improve the training stability of the reward fine-tuning stage of LaSRO.
The terms $\mathcal{S}[\mathcal{R}_{\psi}(\mathbf{z}_*, \mathbf{c})]$ from Eq. \ref{eq:ft} are defined as:
\begin{equation}
    \mathcal{S}[\mathcal{R}_{\psi}(\mathbf{z}_1, \mathbf{c})] = \mathrm{clip}\Big( \frac{\mathcal{R}_{\psi}(\mathbf{z}_1, \mathbf{c}) - \bar{\mathcal{R}}_{\psi}(\mathbf{z}_1, \mathbf{c})}{\mathcal{R}^{\text{90\%}}_{\psi}(\mathbf{z}_1, \mathbf{c})}, -\inf, 1\Big)
\end{equation}
\begin{equation}
    \mathcal{S}[\mathcal{R}_{\psi}(\mathbf{z}_2, \mathbf{c})] = \mathrm{clip}\Big( \frac{\mathcal{R}_{\psi}(\mathbf{z}_2, \mathbf{c}) - \bar{\mathcal{R}}_{\psi}(\mathbf{z}_2, \mathbf{c})}{\mathcal{R}^{\text{90\%}}_{\psi}(\mathbf{z}_2, \mathbf{c})}, -\inf, 1\Big)
\end{equation}
Namely, we keep track of the moving average of $\mathcal{R}_{\psi}(\mathbf{z}_*, \mathbf{c})$
as $\bar{\mathcal{R}}_{\psi}(\mathbf{z}_*, \mathbf{c})$,
and we further keep track of the moving $90-$percentile of $\mathcal{R}_{\psi}(\mathbf{z}_*, \mathbf{c}) - \bar{\mathcal{R}}_{\psi}(\mathbf{z}_*, \mathbf{c})$ as $\mathcal{R}^{\text{90\%}}_{\psi}(\mathbf{z}_*, \mathbf{c}) > 0$.
After normalizing $\mathcal{R}_{\psi}(\mathbf{z}_*, \mathbf{c})$, we clip its maximum value to be $1$ but not its minimum value.
We find $\mathcal{S}$ easy to implement, effectively stabilizing reward fine-tuning, and easing hyper-parameter tuning regarding the learning rates and the loss coefficients in Eq. \ref{eq:ft}.

\section{Connection between LaSRO and Value-based RL} \label{app:value-based-rl}

In Sec. \ref{sec:insight}, we argue that LaSRO shares several functional and structural similarities with value-based RL methods.
Here we provide more details of this comparison.

\paragraph{Details of comparison (i)}
The TD learning loss $\mathcal{L}_{\text{TD}}$ for value-based RL with an MDP of horizon $H=2$ is largely equivalent to the surrogate reward loss $\mathcal{L}_{\text{surr}}$ from Eq. \ref{eq:pretrain}.
Specifically, when adapted to reward fine-tuning two-step DMs with $\mathbf{a}_t = \mathbf{x}_{\tau_{t+1}} = \mathbf{z}_{t+1} + Z$ (the MDP notation from Sec. \ref{sec:mdp} and the notation of $\mathbf{z}_t$ from Sec. \ref{sec:pretrain}), we have
\begin{equation}
    \mathcal{L}_{\text{TD}}(\psi) =
    \mathbb{E}_{(\mathbf{x}_{\tau_t}, \mathbf{a}_t, \mathbf{x}_{\tau_{t+1}}, \mathbf{c})} \big[ \big( Q_{\psi}(\mathbf{x}_{\tau_t}, \mathbf{a}_{t}, \mathbf{c}) - \big(r + \gamma\max_{\mathbf{a}_{t+1}} Q_{\psi}(\mathbf{x}_{\tau_{t+1}}, \mathbf{a}_{t+1}, \mathbf{c})\big) \big)^2 \big] 
\end{equation}
where $r$ is an image-space reward signal.
Since $Q_{\psi}(\mathbf{x}_{\tau_t}, \mathbf{a}_t, \mathbf{c}) = Q_{\psi}(\mathbf{z}_t, \mathbf{z}_{t+1}+Z, \mathbf{c})$ evaluates the ``quality'' of $\mathbf{z}_{t+1}$, it is functionally equivalent to $\mathcal{R}_{\psi}(\mathbf{z}_{t+1}, \mathbf{c})$.
Consequently, we have
\begin{equation}
        \mathcal{L}_{\text{TD}}(\psi) = 
         \mathbb{E} \big[ \big( \mathcal{R}_{\psi}(\mathbf{z}_{t+1}, \mathbf{c}) - \big(r(\mathcal{D}(\mathbf{z}_{t+1}), \mathbf{c}) + \gamma\max_{\mathbf{z}_{t+2}} \mathcal{R}_{\psi}( \mathbf{z}_{t+2}, \mathbf{c})\big) \big)^2 \big]
\end{equation}
Given that $H=2$, and set the discount factor $\gamma =0$ (we equally care about the individual reward of one-step and two-step LCM results), we further have 
\begin{equation}
    \mathcal{L}_{\text{TD}}(\psi) = \mathbb{E}_{ (\mathbf{z}_{1}, \mathbf{c})} \big[ \big(\mathcal{R}_{\psi}( \mathbf{z}_{1}, \mathbf{c}) - r(\mathcal{D}(\mathbf{z}_{1}), \mathbf{c}) \big)^2 \big] + \mathbb{E}_{ (\mathbf{z}_{2}, \mathbf{c})} \big[ \big(\mathcal{R}_{\psi}( \mathbf{z}_{2}, \mathbf{c}) - r(\mathcal{D}(\mathbf{z}_{2}), \mathbf{c}) \big)^2 \big]
\end{equation}
which is essentially the surrogate reward loss $\mathcal{L}_{\text{surr}}$ in an L2 regression form (instead of the binary classification form).

\paragraph{Details of comparison (iii)}
Similar to value-based methods which usually feature more efficient off-policy sampling, LaSRO explores the image (more precisely the latent) space in an off-policy manner.
On-policy sampling refers to sampling data strictly according to the current RL policy, i.e. $p_{\theta}(\mathbf{x}_{\tau_H}|\mathbf{x}_{\tau_0}, \mathbf{c})$, for reward fine-tuning two-step DMs with $H=2$.
Exploration is limited in this case, since given the \textit{fixed} initial noise and prompt $(\mathbf{x}_{\tau_0}, \mathbf{c})$, the stochasticity of the action distribution only comes from one step of noise injection. 
This limitation hinders most policy gradient methods, which require on-policy samples due to the application of the policy gradient theorem.
In contrast, value-based RL usually leverages off-policy samples from arbitrary probability distributions to learn value functions.
In both training stages of LaSRO, we explore the latent space by sampling both the initial noise $\mathbf{x}_{\tau_0}$ and the injected noise $Z$, rather than restricted to a fixed initial noise (i.e., on-policy samples). 
This off-policy sampling helps LaSRO to achieve more effective exploration.

\section{Implementation Details} \label{app:impl}

\paragraph{T2I-prompt-873K}
We curate a set of 873k unique prompts from the text prompts of a combination of DiffusionDB \cite{wang2022diffusiondb} and open-prompts (\href{https://github.com/krea-ai/open-prompts}{link}).
Specifically, we first combine the two prompt sets to obtain a set of 2.9M prompts.
After post-processing to de-duplication, we obtain 1.5M unique prompts.
Then we adopt the active prompt selection from UniFL \cite{zhang2024unifl} to create a more diverse and balanced prompt subset where (almost) all prompt has its closest prompt at least greater than some thresh $T$ for some distance measurement. 
We measure the distance between prompts as the normalized L2 distance of text features from \href{https://huggingface.co/sentence-transformers/all-mpnet-base-v2}{this} text encoder and use \href{https://faiss.ai/index.html}{FAISS} for nearest neighbor search.
We finally ended up with 873k prompts.

\paragraph{Pre-trained LCM based on SSD-1B}
To perform latent consistency distillation using SSD-1B \cite{gupta2024progressive} as the teacher, we generate 700k (prompt, image) pairs as training data from 700k prompts sampled from T2I-prompt-873k.
Each image is of resolution $1024^2$ and generated by SDXL for 40 steps and SDXL-refiner (added noise of strength $0.2$) for another 8 steps with a guidance scale of $7.0$.
We train this LCM with most hyper-parameters similar to the original LCM paper, with the exception that we use $100$ sampling steps instead of $50$ (i.e., a skipping step of $10$ instead of $20$).
We use a fixed learning rate of $3e-6$ with a batch size of $128$ for 150k steps, resulting in better $\le2$ step image generation than in the original paper \cite{luo2023latent}.
The inference time for this 2-step LCM for a $1024^2$ image is around $0.15$ seconds on an A100.

\paragraph{Architecture of Surrogate Reward}
Regarding the architecture of the surrogate reward, we use the UNet's down- and mid-block of the pre-trained \href{https://huggingface.co/segmind/Segmind-Vega}{Segmind-Vega} (a diffusion model distilled from SDXL) as the backbone and use stacks of (4 by 4 convolution with stride 2, group normalization \cite{wu2018group} with 32 groups, SiLU activation \cite{elfwing2018sigmoid}), similar to SDXL-Lightning \cite{lin2024sdxl}, and a linear layer on top of them to produce the final scalar output of $\mathcal{R}_{\psi}(\mathbf{z}, \mathbf{c})$.
During both the pre-training stage and the fine-tuning stage, we update all parameters involved in the surrogate model (no frozen weights).

\paragraph{Pre-training Stage of LaSRO}
The hyper-parameter $N_s$ reflects how hard the exploration problem is for obtaining W/L samples of differed reward values (e.g., the Attribute Binding Score (\%) is rather discrete).
We set $N_s = 1$ when optimizing Image Reward and $N_s=6$ otherwise.
For both one-step and two-step results $\{\mathbf{z}_1\}$ and $\{\mathbf{z}_2\}$, we select the winning and losing (W/L) pair by ranking all $z_*$ within each of the two groups and selecting the best and the worse according to the target reward signal $r$.
This gives us $(\mathbf{z}_1^w, \mathbf{z}_1^l, \mathbf{c})$ and $(\mathbf{z}_2^w, \mathbf{z}_2^l, \mathbf{c})$ that have the maximum reward discrepancy between $\mathbf{z}_*^w$ and $\mathbf{z}_*^l$, facilitating surrogate reward learning.
The sampling strategy for prompts and injected noise is random sampling.
We found it not a critical design choice as long as we vary the initial noise (off-policy exploration).
In the pre-training stage for optimizing the Image Reward, we train the surrogate reward for 30k steps. 
In the pre-training stage for the Attribute Binding Score and the Text Alignment Score, we further fine-tune this trained version for an additional 2k and 10k steps, respectively.
By leveraging the surrogate reward learned from Image Reward, we accelerate the pre-training stage for other reward signals as Image Reward is a rather general image metric.
We use a fixed learning rate of $\eta=1e-5$ for all three metrics.
The training time of the pre-training stage of the surrogate reward for Image Reward is around 12 hours on a single node of 8 H100 GPUs.

\paragraph{Fine-tuning Stage of LaSRO}
In the fine-tuning stage, we use the same setup of $N_s$ and W/L pair-finding mechanism in the pre-training stage.
During the online adaptation of the surrogate reward, specifically in line 6 from Alg. \ref{algo:ft}, we randomly select one $\mathbf{z}_1$ out of $N_s$ sampled $\{\mathbf{z}_1\}$ and similarly for $\mathbf{z}_2$ to compute $\mathcal{S}[\mathcal{R}_{\psi}(\mathbf{z}_*, \mathbf{c})]$, as we find it more efficient.
We use the original loss $\mathcal{L}_{lcm}$ used to train the LCM and the original training data (700k text-image pairs as introduced previously) as a regularizer during the fine-tuning stage.
When optimizing the Image Reward, we alternate between updating the diffusion model for $N_1=1$ step and adapting the surrogate reward for $N_2=1$ step.
When optimizing the Attribute Binding score, we set $N_1=10$ and $N_2=20$.
For the Text Alignment Score, it is $N_1=10$ and $N_2=50$.
The normalization and clipping function $\mathcal{S}$ applied to the reward optimization loss makes hyper-parameter tuning more robust. 
We set $c=500, c_1=0.5, c_2=1.0$ and the learning rate $\eta_1=3e-6, \eta_2=3e-8$ when optimizing all three rewards.
For RL methods compared including LaSRO, we report the results using checkpoints obtained with EMA update of a rate $\mu=0.95$ to ensure more stable performance across iterations.
The training time of the fine-tuning stage of LaSRO for Image Reward is around 48 hours on a single node of 8 H100 GPUs.

\paragraph{GORS-LCM}
GORS-LCM is an RL method based on GORS \cite{huang2023t2i}, which follows a simple generate-filter-distill paradigm and is adapted to fine-tuning LCMs.
It shares a similar form as the original LCM learning process.
In each iteration, it samples $N_s=6$ times from the two-step LCM to obtain $\{\mathbf{z}_1\}$ and $\{\mathbf{z}_2\}$ given prompt $\mathbf{c}$, selects the best $\mathbf{z}_1^{\text{best}}$ and $\mathbf{z}_2^{\text{best}}$ from them according to the target reward $r$, and then trains with the LCM loss $\mathcal{L}_{\text{lcm}}$ to distill from SSD-1B using data $(\mathbf{z}_*^{\text{best}}, \mathbf{c})$ generated online instead of the offline distillation dataset (e.g., the 700k pairs described previously).
While converging slowly due to no reward gradient guidance, this method is stable and can serve as an RL baseline for arbitrary rewards.
We further apply regularization via $\mathcal{L}_{\text{lcm}}$ with the 700k (text, image) pairs, similar to LaSRO.

\paragraph{RLCM/DDPO}

RLCM \cite{oertell2024rl} is a policy-based RL method that adapts DDPO \cite{black2023training} to LCMs, essentially applying PPO \cite{schulman2017proximal} to perform reward fine-tuning.
We adopt most of the hyper-parameters of RLCM (for its case of optimizing the aesthetics score), except that we use a batch size of $128$ and a reduced learning rate of $1e-5$ to ensure training stability.
Moreover, we find that using LoRA updates \cite{hu2021lora} instead of full LCM weight updates helps to improve convergence as LoRA can be seen as a form of regularization through model parameter preservation \cite{choi2024simple}.
To further mitigate the training instability caused by RLCM/DDPO, we carefully turn the hyper-parameters to apply the regularization loss $\mathcal{L}_{\text{lcm}}$ with the 700k (text, image) data pairs in (similar to LaSRO).

\paragraph{PSO/Diff-DPO}

PSO \cite{miao2024tuning}, particularly its online version, is an online variant of Diffusion-DPO \cite{wallace2024diffusion} adapted to step-distilled diffusion models, such as LCMs.
We largely follow the hyper-parameters of Diffusion-DPO and the online PSO when reward fine-tuning the two-step LCMs (with a batch size of $128$).
Similar to our experiments of RLCM/DDPO, we find that updating LCM weights using LoRA serves as a regularization technique that leads to improvements in both the numerical and visual results (in contrast, LaSRO works well with or without LoRA).
Similar to other RL baselines, we apply regularization via $\mathcal{L}_{\text{lcm}}$ for a fair comparison to LaSRO.
Note that, PRDP \cite{deng2024prdp} can be seen as another online variant of Diffusion-DPO that applies reward-weighted regression in a reward-adapted manner.
The difficulties encountered in reward fine-tuning \(\leq 2\)-step diffusion models using PSO/Diffusion-DPO are also relevant to PRDP.

\paragraph{Ablation Study (a)}
We replace our latent-space surrogate reward design based on pre-trained diffusion models with CLIP \cite{radford2021learning} or BLIP \cite{li2022blip} to validate our choices.
Specifically, we use the \texttt{CLIP-ViT-L/14} from the original CLIP paper and a similar BLIP model fine-tuned for VQA tasks (ViT-L for the vision backbone).
We train both in the latent space of SDXL as a fair comparison with our design in LaSRO.
Both have a similar number of model parameters as the pre-trained UNet that we use.
Both are trained in the pre-training and the reward fine-tuning (online adaptation) stages.

\paragraph{Ablation Study (b)}
In this ablation study, we examine the alternative reward fine-tuning scheme adopted by most existing fine-tuning methods with direct reward back-propagation, including RG-LCD \cite{li2024reward}.
This alternative does not target 2-step image generation. 
Instead, in each reward fine-tuning step, it randomly chooses a timestep $t$, which refers to the notation used in the consistency distillation loss $\mathcal{L}_{\text{lcm}}$ from Eq. \ref{eq:lcm_loss}. 
Specifically, $t\in \{1, ..., 50\}$ in the original LCM paper with a skipping step of $20$ and $t\in \{1, ..., 100\}$ in our LCM training setup with a skipping step of $10$.
For this timestep $t$, which could be anywhere in a denoising trajectory, the alternative scheme optimizes $\mathcal{R}_{\psi}(f_{\theta}(\mathbf{x}_{t}, t, \mathbf{c}), \mathbf{c})$ with $\mathbf{x}_{t}$ generated by adding Gaussian noise $Z_t$ of level $t$ to the samples from the 700k (text, image) training pairs.
In comparison, LaSRO optimizes $\mathcal{R}_{\psi}(\mathbf{z}_*, \mathbf{c})$, where $\mathbf{z}$ is generated completely online, either from the $1^{\text{st}}$-step LCM results given random initial noise, or from the $2^{\text{nd}}$-step LCM results given the $1^{\text{st}}$-step results.
This leads to improved effectiveness in fine-tuning $\le2$-step image generation.

\section{Additional Experiments Results and Implementation Details} \label{app:overoptimize}

\paragraph{Reward over-optimization}
Reward over-optimization \cite{skalse2022defining} is a common issue for reward fine-tuning diffusion models, referring to optimizing overly on an imperfect reward so that the model performance measured by some unknown ``true'' objective gets dropped.
Visually this usually means blurry images or abnormal image compositions and visual patterns where the fine-tuned model no longer preserves the high fidelity of its outputs (e.g., the Fig. 5 in the DDPO paper \cite{black2023training}).
While no practical techniques are $100\%$ exempt from reward over-optimization, we adopt several to mitigate this issue.
Namely, for all compared methods, we use the latent consistency distillation loss $\mathcal{L}_{\text{lcm}}$ as a regularizer on a large distillation dataset (T2I-prompt-873k and the corresponding images), which essentially penalizes large KL divergence between the fine-tuned model and the reference one (e.g., SSD-1B).
When optimizing the Image Reward \cite{xu2024imagereward}, we use a large number of diverse prompts from the same T2I-prompt-873k set, which is generated via the active prompt selection that helps mitigate reward over-optimization \cite{zhang2024unifl}.
We carefully tune the learning rate (in most cases to be lowered) of all methods with qualitative evaluation to help monitor the image fidelity of the fine-tuned model's outputs.
While one can resort to FID \cite{heusel2017gans} to quantitatively evaluate the image fidelity (whose results will be reported shortly), it requires a reference image set and the scores do not always align with human judgment.

\begin{figure}[h]
  \centering
    \includegraphics[width=0.65\textwidth]{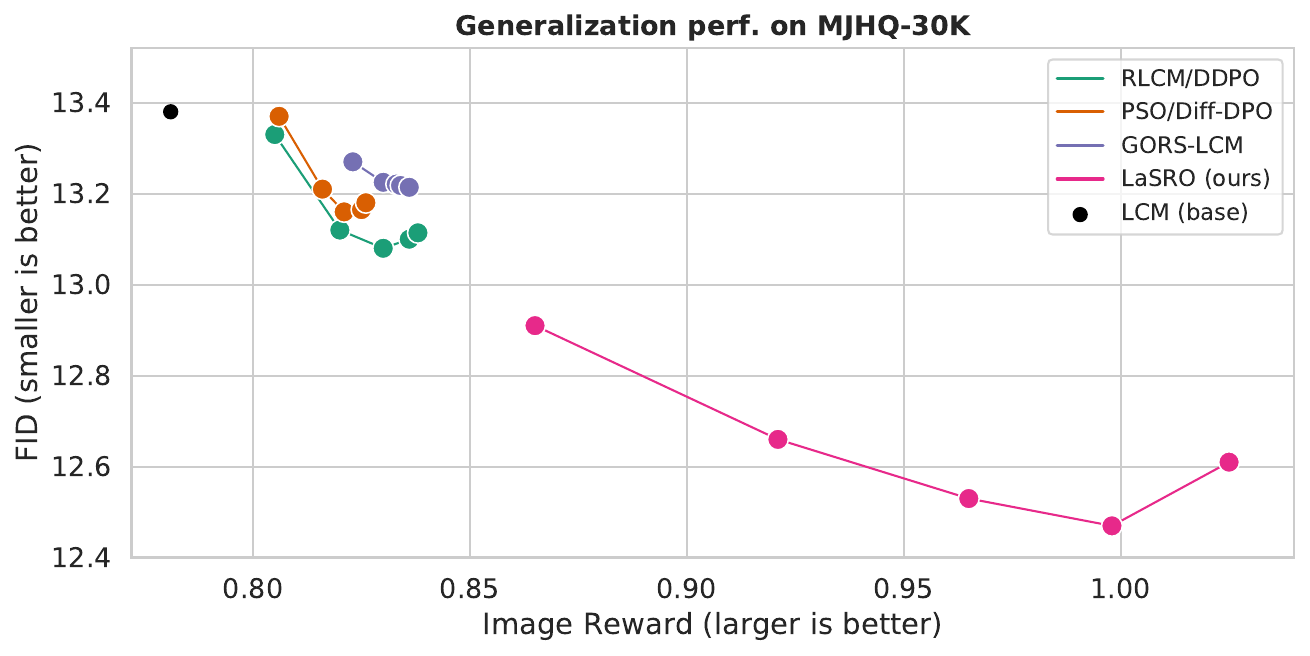}
  \caption{
  The curves that present the tradeoff between FID and Image Reward for the LCM baseline and the fine-tuned versions with all compared methods from Fig. \ref{fig:plots_1}. Each point is generated by one of the checkpoints with $1000, 2000, 3000, 4000$, or $5000$ reward fine-tuning update steps and measured on two-step generated images of $1024^2$ pixels. The closer a point is to the lower right corner, the better the overall model performance is. LaSRO consistently outperforms other RL baselines.}
  \label{fig:tradeoff}
\end{figure}

\paragraph{Tradeoff - FID vs. Image Reward}
The inherent trade-off between preserving high fidelity and achieving target optimization is manifested with the tradeoff between FID vs. Image Reward.
We use FID \cite{heusel2017gans}, a prompt-free metric, to measure the image fidelity of the results from fine-tuned diffusion models.
We measure it between the generated images from prompts in MJHQ-30K and the human-curated reference image set provided by MJHQ-30K.
Upon aggressively optimizing the Image Reward score using LaSRO, we report curves of FID vs. Image Reward in Fig. \ref{fig:tradeoff} of two-step LCMs for all compared methods in Fig. \ref{fig:plots_1}.
Note that, in our experiments, we evaluate LaSRO on a well-trained distilled DM baseline to see if it can still provide meaningful improvement without losing visual qualities (reward hacking).
The results further demonstrate the superior performance of LaSRO.

\paragraph{Implmentation Details for Tab. \ref{tab:numerical}} \label{app:numerical}
Given FID as an auxiliary metric, we can summarize the Image Reward performance of each method (i.e., the curves shown in Fig. \ref{fig:plots_1}) as a single numerical number by reporting the result of the checkpoint with the lowest FID (the elbow rule), which is supposedly better than that of the LCM baseline.
We report the numerical results in Tab. \ref{tab:numerical}, where we also show the Image Reward of other step-distilled diffusion model baselines including SDXL-Lightning \cite{lin2024sdxl} (2-step) and SDXL-Turbo 
\cite{sauer2023adversarial} (1- and 2-step), both based on adversarial distillation.
Note that, SDXL-Turbo mainly focuses on image generation of $512^2$ pixels and SDXL-Lightning only supports $\ge 2$ step generation.
In addition, we examine applying LaSRO toward improving SDXL-Turbo with Image Reward, whose result showcase that LaSRO is not specific to LCMs. 
Namely, we do not use $\mathcal{L}_{\text{lcm}}$ as regularization (the pre-trained discriminator of SDXL-Turbo is not publicly released, and thus we can not use its original adversarial distillation loss for regularization) and reduce the learning rate $\eta_1=1e-8$ in the fine-tuning stage to mitigate reward over-optimization.

\section{Additional Visual Comparison (see next page)}  \label{app:visual}


\begin{figure}[h]
  \centering
    \includegraphics[width=0.97\textwidth]{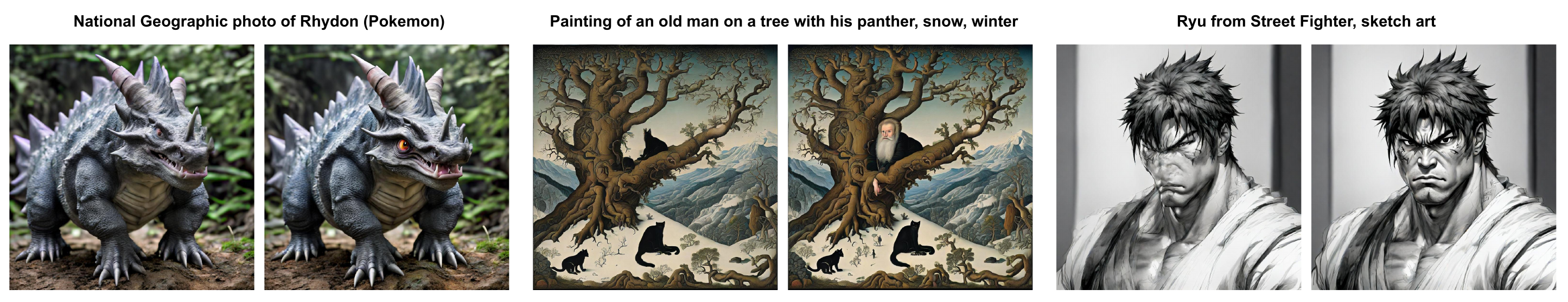}
    \includegraphics[width=0.97\textwidth]{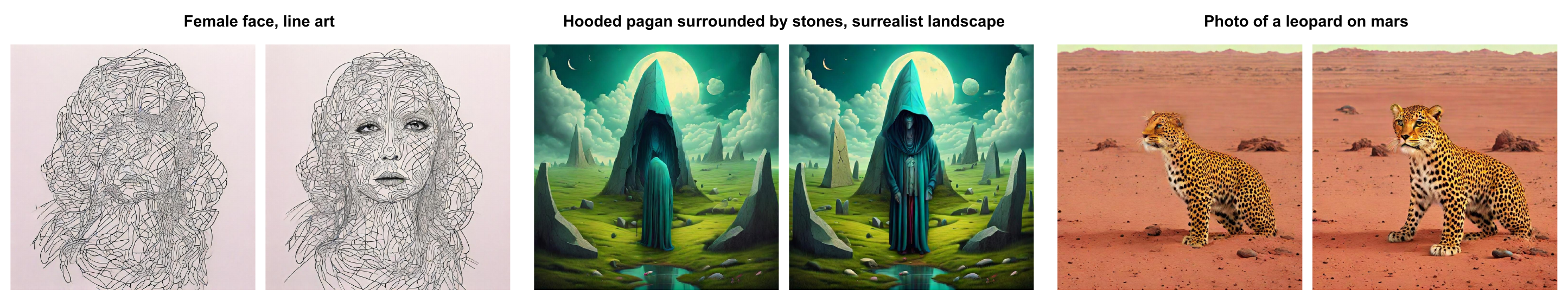}
    \includegraphics[width=0.97\textwidth]{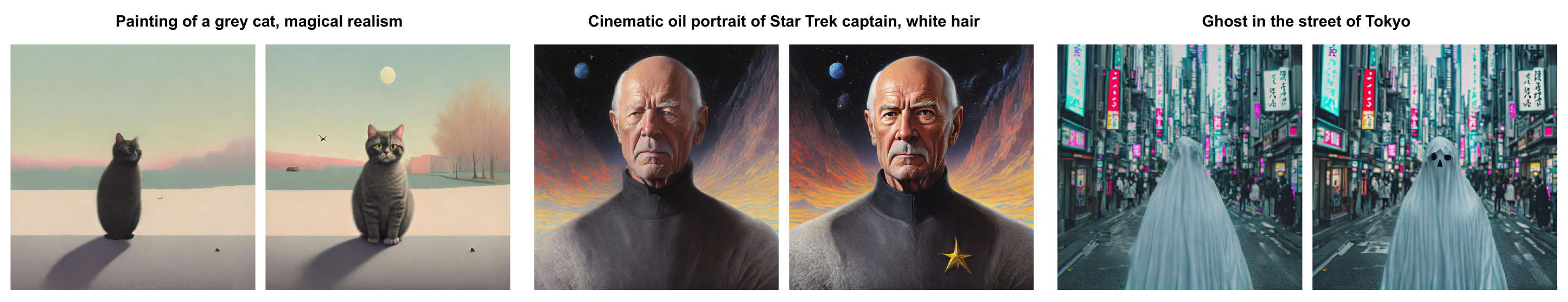}
    \includegraphics[width=0.97\textwidth]{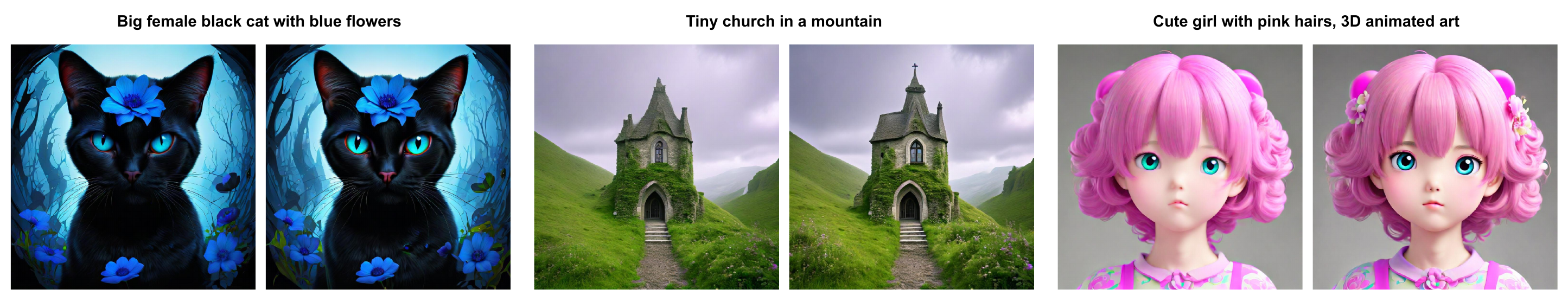}
    \includegraphics[width=0.97\textwidth]{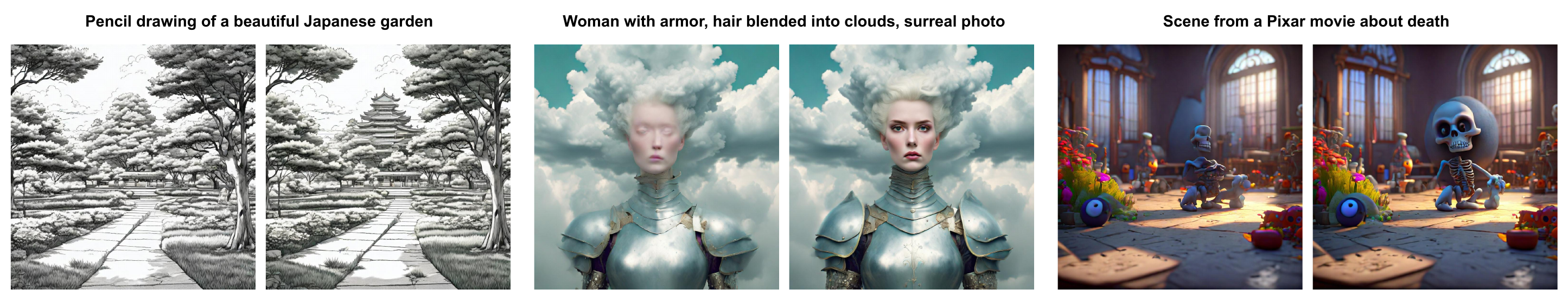}
    \includegraphics[width=0.97\textwidth]{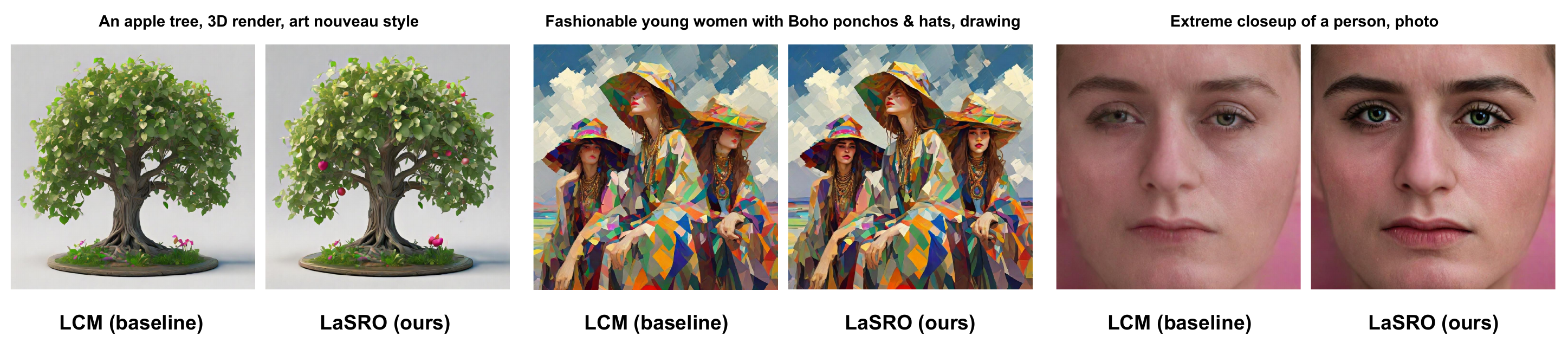}
  \caption{
  Additional visual comparison of two-step LCM results from the LCM baseline and the fine-tuned LCM via LaSRO with Image Reward. All images are of $1024^2$ resolution with prompts randomly sampled from T2I-prompt-873K}
  \label{fig:many_pics}
\end{figure}


\end{document}